\documentclass[nonacm]{acmart}
\AtBeginDocument{%
  }

\setcopyright{none} %
\copyrightyear{2023}
\acmYear{2023}
\acmDOI{XXXXXXX.XXXXXXX}

\usepackage{hyperref}
\usepackage{url}
\usepackage{booktabs}       %
\usepackage{amsfonts}       %
\usepackage{nicefrac}       %
\usepackage{microtype}      %
\usepackage{xcolor}         %
\usepackage{natbib}
\usepackage{multirow}
\usepackage{graphicx}
\graphicspath{ {./images/} }
\usepackage{subcaption}
\usepackage{empheq}
\usepackage{colortbl}
\usepackage{color}
\definecolor{gray}{rgb}{0.2,0.2,0.3}

\usepackage{amsmath}
\usepackage{makecell}
\DeclareMathOperator*{\argmax}{arg\,max}

\usepackage{dsfont}
\usepackage{tikz}
\usepackage{xspace}
\usepackage{tabularx}
\newcommand{\ie}{\textit{i.e.}\xspace}
\newcommand{\eg}{\textit{e.g.}\xspace}
\usepackage{algorithm}
\usepackage[noend]{algpseudocode}
\usepackage{bbm}
\usepackage{numprint}
\usepackage{cleveref}
\usepackage{wrapfig}
\usepackage{pifont}
\usepackage{circledsteps}
\usepackage{xspace}
\usepackage{tabularx}
\usepackage{makecell}
\usepackage{todonotes}
\usepackage{enotez}
\usepackage{setspace}

\newif\ifdraft

\draftfalse

\ifdraft
    \newcommand{\my}[1]{\textcolor{teal}{m: #1}}
    
    \newcommand{\ftd}[1]{\todo{\textcolor{violet}{franzi: #1}}}
    \newcommand{\franzi}[1]{\textcolor{violet}{franzi: #1}}
    \newcommand{\ilia}[1]{\textcolor{green}{ilia: #1}}
    \newcommand{\anvith}[1]{\textcolor{blue}{2D: #1}}
    \newcommand{\sierra}[1]{\textcolor{blue}{sierra: #1}}
    \newcommand{\jonas}[1]{\textcolor{olive}{jonas:#1}}
    \newcommand{\patty}[1]{\textcolor{cyan}{patty:#1}}
    \newcommand{\varun}[1]{\textcolor{red}{VC:#1}}
    \newcommand{\nicolas}[1]{\textcolor{olive}{NP:#1}}
    \newcommand{\stephan}[1]{\textcolor{orange}{stephan: #1}}
    
    \newcommand{\ahmad}[1]{\textcolor{magenta}{ahmad:#1}}

\else
    \newcommand{\my}[1]{}
    \newcommand{\franzi}[1]{}
    \newcommand{\ilia}[1]{}
    \newcommand{\anvith}[1]{}
    \newcommand{\sierra}[1]{}
    \newcommand{\jonas}[1]{}
    \newcommand{\patty}[1]{}
    \newcommand{\varun}[1]{}
    \newcommand{\nicolas}[1]{}
    \newcommand{\stephan}[1]{}
    \newcommand{\daivd}[1]{}
    \newcommand{\ahmad}[1]{}
    \newcommand{\ftd}[1]{}
\fi

\makeatletter
\newcommand{\rpriv}[1][\@nil]{%
    \def\tmp{#1}%
    \ifx\tmp\@nnil
        \rho_\text{priv}
    \else
        #1
    \fi}

\newcommand{\rfair}[1][\@nil]{%
    \def\tmp{#1}%
    \ifx\tmp\@nnil
        \gamma
    \else
        #1
    \fi}

\newcommand{\tpriv}[1][\@nil]{%
    \def\tmp{#1}%
    \ifx\tmp\@nnil
        \tau_\text{priv}
    \else
        #1
    \fi}

\newcommand{\tfair}[1][\@nil]{%
    \def\tmp{#1}%
    \ifx\tmp\@nnil
        \tau_\text{build}
    \else
        #1
    \fi}

\makeatother

\newcommand{\model}{\theta}

\newcommand{\eps}{\varepsilon}

\algnewcommand\algorithmicinput{\textbf{Input:}}
\algnewcommand\Input{\item[\algorithmicinput]}%

\newcommand{\Dpriv}{D_\text{private}}

\newcommand{\Dpub}{D_\text{public}}

\newcommand{\numTeachers}{B}
\newcommand{\numClasses}{K}

\newcommand{\fdpsgd}{FairDP-SGD\xspace}
\newcommand{\fpate}{FairPATE\xspace}
\newcommand{\pf}{Pareto frontier\xspace}
\newcommand{\pfs}{Pareto frontiers\xspace}
\newcommand{\dpsgd}{DP-SGD\xspace}
\newcommand{\pe}{Pareto-efficient\xspace}
\newcommand{\pie}{Pareto-inefficient\xspace}
\newcommand{\ds}{label shift\xspace}

\pgfkeys{/csteps/fill color=black}
\pgfkeys{/csteps/inner color=white}
\newcommand{\one}{\Circled{\textbf{1}}}
\newcommand{\two}{\Circled{\textbf{2}}}
\newcommand{\three}{\Circled{\textbf{3}}}
\newcommand{\four}{\Circled{\textbf{4}}}
\newcommand{\five}{\Circled{\textbf{5}}}
\newcommand{\six}{\Circled{\textbf{6}}}
\newcommand{\seven}{\Circled{\textbf{7}}}

\newcommand{\eight}{
\hspace{-1.5mm}
\begin{tikzpicture}[baseline=-1mm]
\node[circle, draw, text=black, fill=black, inner sep=1pt, pattern=north east lines, fill opacity=0.3, text opacity=1.0, font=\bfseries] {8};
\end{tikzpicture}
\hspace{-1.5mm}
}
\newcommand{\nine}{
\hspace{-1.5mm}
\begin{tikzpicture}[baseline=-1mm]
\node[circle, draw, text=black, fill=black, inner sep=1pt, pattern=north east lines, fill opacity=0.3, text opacity=1.0, font=\bfseries] {9};
\end{tikzpicture}
\hspace{-1.5mm}
}

\newtheorem{prop}{Proposition}
\newtheorem{definition}{Definition}

\usepackage{ifthen}
\usepackage{tikz}
\usetikzlibrary{arrows.meta}
\usetikzlibrary{calc, babel, decorations.text}
\usetikzlibrary{positioning}
\usepgflibrary{shapes.arrows}
\usetikzlibrary{patterns}
\usetikzlibrary{decorations.text}
\usetikzlibrary{intersections} 
\usetikzlibrary{arrows, fit}
\usetikzlibrary{3d}

\usepackage{pgfplots}
\pgfplotsset{compat=1.8}
\usepgfplotslibrary{colorbrewer}
\usepgfplotslibrary{colormaps}

\makeatletter
\tikzset{arc style/.initial={}}
\pgfdeclareshape{half circle}{
    \inheritsavedanchors[from=circle]
    \inheritanchorborder[from=circle]

    \inheritanchor[from=circle]{center}
    \inheritanchor[from=circle]{south}
    \inheritanchor[from=circle]{west}
    \inheritanchor[from=circle]{north}
    \inheritanchor[from=circle]{east}

    \inheritbackgroundpath[from=circle]

    \beforebackgroundpath{
        \pgfkeys{/tikz/arc style/.get=\tmp}
        \expandafter\tikzset\expandafter{\tmp}
        \tikz@options

        \radius \pgf@xa=\pgf@x
        \centerpoint \pgf@xb=\pgf@x \pgf@yb=\pgf@y

        \advance\pgf@yb by \pgf@xa
        \pgfpathmoveto{\pgfpoint{\pgf@xb}{\pgf@yb}}
        \pgfpatharc{90}{-90}{\pgf@xa}

        \pgfusepath{fill}
    }
}
\makeatother

\tikzset{
    partial ellipse/.style args={#1:#2:#3}{
            insert path={+ (#1:#3) arc (#1:#2:#3)}
        }
}

\pgfdeclarepatternformonly{super ultra thick north east lines}{\pgfqpoint{-2pt}{-2pt}}{\pgfqpoint{10pt}{10pt}}{\pgfqpoint{10pt}{10pt}}%
{
  \pgfsetlinewidth{2pt}
  \pgfpathmoveto{\pgfqpoint{-1pt}{-1pt}}
  \pgfpathlineto{\pgfqpoint{9pt}{9pt}}
  \pgfusepath{stroke}
}

\begin{document}

\title{Learning with Impartiality to Walk on the Pareto Frontier of Fairness, Privacy, and Utility}

\author{Mohammad Yaghini}
\email{mohammad.yaghini@mail.utoronto.ca}

\author{Patty Liu}
\email{patty.liu@mail.utoronto.ca}

\author{Franziska Boenisch}
\email{franziska.boenisch@vectorinstitute.ai}

\author{Nicolas Papernot}
\email{nicolas.papernot@utoronto.ca}

\affiliation{%
  \institution{University of Toronto \& Vector Institute}
  \city{Toronto}
  \state{Ontario}
  \country{Canada}
}

\begin{abstract}
Deploying machine learning (ML) models often requires both fairness and privacy guarantees.
Both of these objectives present unique trade-offs with the utility (e.g., accuracy) of the model.
However, the mutual interactions between fairness, privacy, and utility are less
well-understood. 
As a result, often only one objective is optimized, while the others are tuned as hyper-parameters. 
Because they implicitly prioritize certain objectives, such designs bias the model in pernicious, undetectable ways.
To address this, we adopt impartiality as a principle: design of ML pipelines should not favor one objective over another. 
We propose impartially-specified models, which provide us with accurate \pfs that show the inherent trade-offs between the objectives.
Extending two canonical ML frameworks for privacy-preserving learning, we provide two methods (\textit{\fdpsgd} and \textit{\fpate}) to train impartially-specified models and recover the \pf. Through theoretical privacy analysis and a comprehensive empirical study,
we provide an answer to the question of where fairness mitigation should be integrated within a privacy-aware ML pipeline.

\end{abstract}

\maketitle

\section{Introduction}
\label{sec:intro}

From medical applications~\cite{irvin2019chexpert} to infrastructure planning from census data~\cite{dp_census}, deploying machine learning (ML) models in critical contexts often requires not only utility (accuracy) guarantees, but also fairness and privacy assurances.    
Prior attempts to mitigate the tension between fairness, privacy, and utility either attempt to adapt 
 private learning 
for improved trade-offs with fairness
~\cite{xu2021removing,zhang2021balancing,tran2021fairness}, or integrate privacy constraints into bias mitigation methods~\cite{jagielski2019differentially}.

We argue that such works obscure the complexity of multi-objective decision making under the veil of ``tuning:''
they optimize for one metric while specifying arbitrary limits on others.
This practice raises fundamental issues.
Notably, it puts socially-salient choices at the behest of algorithm designers and engineers. 
This may result in one objective being relegated to secondary consideration---creating potentially dangerous scenarios, such as introducing additional privacy leakage in the attempt to increase model fairness, or degrading fairness by introducing privacy~\cite{chang_privacy_2021, vinith}.

Instead of reducing trade-offs between different objectives to a weighted combined metric, we propose algorithms that provide a more informative primitive; such as a trade-off function, or its more general counterpart, a \textit{\pf}.
Richer trade-off representations provide improvements on multiple levels.
First, on a technical level, we avoid the pitfall of premature aggregation, and can produce solutions at least as good (if not better) than with a combined weighted metric. 
Second, we improve the transparency of critical ML models by exposing their inherent trade-offs to the decision makers. 
At this level, these decision makers, e.g., lawyers, humanities' experts, elected officials, and the public can make informed decisions regarding the appropriate operating point on the \pf. 
Importantly, this does not require them having expertise in optimization or ML. They can use the \pf over different objectives to choose societally favorable, but still technically feasible, specifications. 

\begin{wrapfigure}[16]{r}{0.5\textwidth}
     \includegraphics[width=0.49\textwidth]{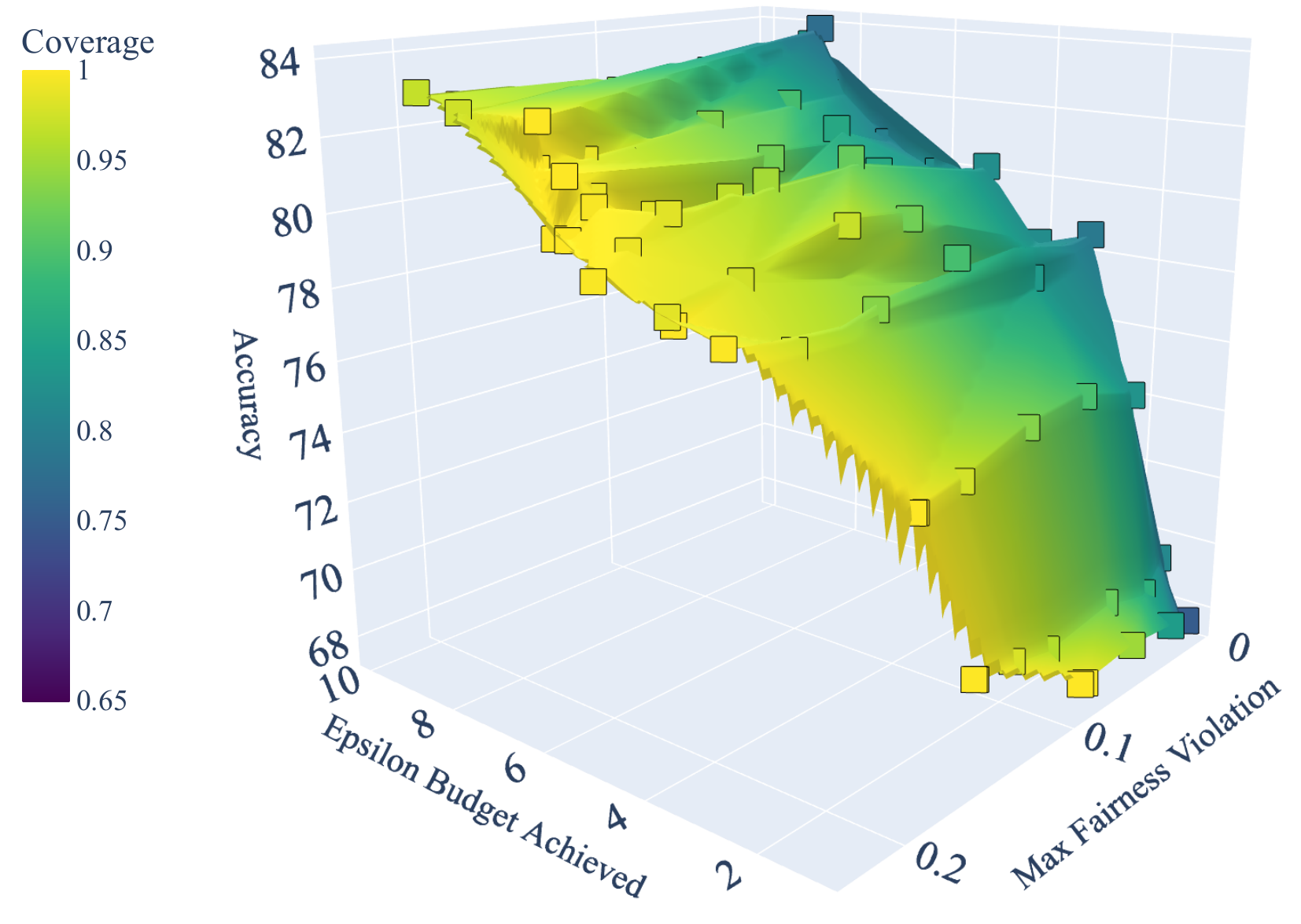}
     \label{fig:pareto-example}
     \caption{Percentage of answered queries (coverage) as a function of the maximum privacy budget $\varepsilon$ and maximum fairness violation $\gamma$ achieved during training with our \fpate.
    }
\end{wrapfigure}

Let us illustrate these improvements with \Cref{fig:pareto-example}. We showcase a typical \pf calculated using one of our approaches
(\Cref{sec:fairpate}) on the UTKFace dataset for a binary prediction task which uses a demographic parity fairness mitigation. 
Consider two scenarios. 
In scenario~1, given a differential privacy budget of $\varepsilon=3$, a fairness violation of $\gamma = 0.05$ will allow us to answer 84\% of queries posed to the model.  

Now scenario 2, where we relax our fairness violations constraint to $\gamma=0.1$.
This allows us to answer up to 89\% of the queries (+5\% improvement), and improve accuracy by 1.5\%, while maintaining the same privacy guarantee. For one decision maker, the loss in fairness that results in switching from scenario 1 to 2 may be warranted by the gains in other metrics---while this may be completely unacceptable to another decision maker. The choice between these scenarios is not a technical one but may have important societal consequences. Thus, it should be left to the decision maker and not the algorithm designer. 

\textit{But how do we obtain such \pfs?}
We propose to adopt \textit{Impartiality} as a principle. That is, we design ML pipelines that do not favour one objective (\eg, fairness or privacy) to the detriment of another.

We start our pursuit of impartiality by studying the interactions between our notions of privacy (differential privacy~\cite{dwork_dp}) and fairness (demographic parity~\citep{calders2010three}). In particular, we characterize the tensions between mitigations developed for these notions. We observe that certain mitigations (for instance, fairness pre-processing) can lead to degraded privacy guarantees. 
We conclude that ML pipelines naively composed without consideration for these tensions are inherently inefficient. In the spirit of impartial designs, we provide two general fairness mitigation strategies that do not cause additional privacy leakage. 

We then turn our attention to more bespoke privacy-preserving ML frameworks. We start with the Differentially Private Stochastic Gradient Descent (\dpsgd)~\cite{abadi2016deep} and consider how we can integrate a fairness mitigation within it. Our analysis informs our first algorithm contribution, namely, \textit{\fdpsgd} which incorporates the aforementioned fairness mitigations. Next, we consider the Private Aggregation of Teacher Ensembles (PATE)~\cite{papernot2018scalable} which, thanks to its modular design, affords us even more freedom in incorporating a fairness mitigation. A thorough privacy-fairness analysis of our options leads us to our second algorithm contribution, \textit{\fpate}. We evaluate our models against a suite of non-impartial models on multiple datasets. We find that our impartial designs often produce the most efficient results, and therefore, naturally surface the \pf---representing the irreconcilable trade-offs between various trustworthiness objectives.
 In summary, our contributions are as follows:

\begin{enumerate}
    \item We argue for the need for richer representations of the multi-objective ML trustworthiness problem in the form of \pfs. These impartial representations allow us to instantiate the baseline model specification problem, where a regulatory body uses the \pf to provide specifications for trustworthy measures. Our discussion focuses on two measures: demographic parity violations $\gamma$  and a differential privacy bound $\varepsilon$. %
    \item To achieve the \pf, we adopt impartiality between our trustworthiness objectives. As a result, we (a) initiate a study to understand the tensions between demographic parity and differential privacy with a focus on the mutual impact of their mitigations and (b) provide fairness mitigation strategies at zero-cost to privacy. Our analysis includes, to the best of our knowledge, the first privacy analysis of a pre-processed fairness mitigation. We then consider bespoke designs for widely-adopted privacy frameworks. Studying limitations of \dpsgd leads us to develop \fdpsgd. To avoid the limitations of \dpsgd, we consider the PATE framework. After a thorough fairness-privacy analysis we present \fpate. We provide privacy analysis for both our algorithms.
    
    \item We provide interactive\footnote{Available at~\href{https://cleverhans-lab.github.io/impartiality_viz/}{https://cleverhans-lab.github.io/impartiality\_viz/}} \pfs for various vision tasks, including a medical Chest Xray disease diagnosis dataset (CheXpert) with known fairness issues~\cite{banerjee_reading_2022, seyyed-kalantari_chexclusion_2020}. Our extensive empirical evaluation compares  various in- and pre-processing
    mitigation techniques. Our results show the inefficiency of non-impartial models: we improve accuracy up to 5\% through careful privacy budget consumption model of \fpate.
    
    \item We provide domain-specific best-practices. These not only include fairness and privacy hyper-parameter choices but also, for the first time, the choice of privacy framework and the corresponding point in the ML pipeline at which the mitigation should be inserted (e.g., before or during training).
\end{enumerate}

\section{Motivation \& Background}
\label{sec:background}

Acknowledging that machine learning models can pose risks to society---much like pollution-emitting industries pose health risks to the environment---it is a reasonable expectation that a regulatory body should produce technical specifications to curb the societal risks of ML models. This is similar to how the Environment Protection Agency (EPA) in the US, or the European emission standards
specify maximum emission ratings.

\paragraph{Problem Setting}
Our problem setting is that of \textit{baseline specification}; where a regulatory body needs to decide on the (hyper)parameters for trustworthiness guarantees. Since model builders are always going to optimize for accuracy, the focus of the baseline specification should be on other, societally salient, objectives. In this paper, we focus on fairness and privacy. 
In particular, we limit our study to the demographic parity notion of fairness guarantees parameterized by $\gamma \in [0,1]$ which characterizes the maximum tolerable fairness violations.
For privacy, we consider differential privacy, parametrized by the privacy budget $\varepsilon \in (0, +\infty)$. 
Ideally, the regulatory body would use a \pf over the objectives to choose societally favorable, but still technically feasible specifications for their respective parameters.

In the remainder of this section, we provide the necessary background on machine learning, fairness, and privacy. We rigorously define our notions of fairness and privacy in~\Cref{sec:background-fairness} and \Cref{sec:background-privacy}, and provide a formal definition of Pareto efficiency in~\Cref{sec:background-pareto}.

\paragraph{ML Background}
We assume a classification task where a model $\model: \mathcal{X} \times \mathcal{Z} \mapsto \mathcal{K}$ maps the features $(\mathbf{x}, z) \in \mathcal{X} \times \mathcal{Z}$ to a label $y \in \mathcal{K}$, where: $\mathcal{X}$ is the domain of non-sensitive attributes, $\mathcal{Z}$ is the domain of the sensitive attribute (as a categorical variable), and $\mathcal{K}$ is the domain of the output label (also categorical). Without loss of generality, we will assume $\mathcal{Z} = [Z]$ (\ie $\mathcal{Z} = \{1, \dots, Z\})$ and $\mathcal{K} = [K]$.

\subsection{Fairness: Demographic Parity}
\label{sec:background-fairness}
We base our work on the fairness metric of \emph{demographic parity} which requires that ML models produce similar success rates (\ie, rate of predicting a desirable outcome, such as getting a loan) for all sub-populations~\citep{calders2010three}. 

We note that in a multi-class setting (\ie, $K > 2$), and even in the binary-class settings where the problem does not admit a reasonable notion of the ``desirable outcome'', there can be multiple formulations of the notion of demographic parity (\Cref{sec:dem-parity-notions}). We adopt a natural extension of the well-known binary notion that requires equal rates for any class. Let us first define demographic disparity:

The \emph{demographic disparity} $\Gamma(z, k)$ of subgroup $z$ for class $k$ is the difference between the probability of predicting class $k$ for the subgroup $z$ and the probability of the same event for any other subgroup: 
$
    \Gamma(z, k) := 
    \mathbb{P}[\hat{Y} = k \mid Z =  z] 
    - \mathbb{P}[\hat{Y} = k \mid Z \neq z].
$
In practice, we estimate multi-class demographic disparity for class $k$ and subgroup $z$ with:
$
    \widehat{\Gamma}(z, k) := 
    \frac{|\{\hat{Y}=k, Z = z \}|}{|\{Z = z\}|} - \frac{|\{\hat{Y}=k, Z \neq z\}|}{|\{Z \neq z\}|},
$
where $\hat{Y} = \model(\mathbf{x}, z)$. We define demographic \textit{parity} when the worst-case demographic disparity between members and non-members for any subgroup, and for any class is bounded: 
\begin{definition}[$\gamma$-DemParity] For predictions $Y$ with corresponding sensitive attributes $Z$ to satisfy $\gamma$-bounded demographic parity ($\gamma$-DemParity), it must be that for all $z$ in $\mathcal{Z}$ and for all $k$ in $\mathcal{K}$, the demographic disparity is at most $\gamma$: $\Gamma(z, k) \leq \gamma$.
\end{definition}

\subsection{Privacy: Differential Privacy}
\label{sec:background-privacy}
Differential Privacy (DP)~\cite{dwork_algorithmic_2013} formalizes the intuition that no individual data point should significantly impact the results of an analysis ran on a complete dataset.
This allows it to learn properties of the dataset while ensuring individual data points' privacy.
More formally, $(\varepsilon, \delta)$-DP can be expressed as follows:
\begin{definition}[$(\varepsilon, \delta)$-Differential Privacy]
\label{df:Differential Privacy}
Let $\mathcal{M} \colon \mathcal{D}^* \rightarrow \mathcal{R}$ be a randomized algorithm that satisfies $(\varepsilon, \delta)$-DP with $\varepsilon \in \mathbb{R}_+$ and $\delta \in [0, 1]$ if for all neighboring datasets $D \sim D'$, \ie, datasets that differ in only one data point, and for all possible subsets $R \subseteq \mathcal{R}$ of the result space it must hold that
$\mathbb{P}\left[M(D) \in R\right] \leq e^\varepsilon \cdot \mathbb{P}\left[M(D') \in R\right] + \delta \,$.
\end{definition}
The parameter $\varepsilon$ bounds the maximal difference between the analysis results on the neighboring datasets while the second parameter $\delta$ represents a relaxation of the bound by allowing the results to vary more than the factor $e^\varepsilon$. 
Hence, the total privacy loss is bounded by $\varepsilon$ with a probability of at least $1-\delta$~\cite{dwork_algorithmic_2013}.
Note that smaller $\varepsilon$ correspond to better privacy guarantees for the data.

In ML, there exist two main canonical algorithms to implement DP, first the Private Aggregation of Teacher Ensemble (PATE)~\cite{papernot2016semi} an ensemble-based approach for private knowledge transfer, and the Differential Private Stochastic Gradient Descent (\dpsgd)~\cite{abadi2016deep}. We present background on both and extend upon them in \Cref{sec:fairpate} and \Cref{sec:dpsgd}.

\begin{wrapfigure}[13]{r}{0.4\textwidth} 
\vspace{-0.8cm}
\begin{minipage}{0.4\textwidth}
\begin{algorithm}[H]
\caption{
\textbf{-- Confident-GNMax Aggregator (from \cite{papernot2018scalable})} given a query, consensus among teachers is first estimated in a privacy-preserving way to then only reveal confident teacher predictions.}
\label{alg:confident aggregator}
\begin{algorithmic}[1] \Require input $x$, threshold $T$, noise parameters $\sigma_1$ and $\sigma_2$
\If{$\max_j \{\sum_{i \in [\numTeachers]} n_{i,j}(x)\} + \mathcal{N}(0, \sigma_1^2) \geq T$} \label{line:noisy_gnmax}
	\State\Return $\argmax_{j} \{\sum_{i \in [\numTeachers]} n_{i,j}(\mathbf{x})+\mathcal{N}(0,\sigma^2_2)\}$
    \label{line:noisy_argmax}
\Else
	\State\Return $\bot$
\EndIf
\end{algorithmic}
\end{algorithm}
\end{minipage}
\end{wrapfigure}
\paragraph{PATE}
PATE (\Cref{fig:pate}), takes advantage of an unlabeled public data set $\Dpub$ to conserve the privacy of sensitive data $\Dpriv$. 
Therefore, an ensemble of $\numTeachers$ \textit{teacher} models $\{\model_i\}_{i=1}^\numTeachers$ is trained using disjoint subsets of $\Dpriv$ and their knowledge is transferred to a separate \textit{student} model that can be publicly released.
For the knowledge transfer, trained teachers label query data points from $\Dpub$. 
The final label of the query is the majority over the vote counts $N(\mathbf{x}) = [n_{i,j}]_{\numTeachers \times \numClasses}$, where $\numClasses$ is the number of classes.

PATE estimates the privacy cost of answering queries (\ie labeling data) through \textit{teachers consensus} with higher consensus revealing less information about individual teachers, and, thereby, consuming less privacy costs. 
To take advantage of the fact that estimating consensus is less privacy-costly than answering queries, PATE rejects high-cost queries to save on the privacy budget (see \Cref{alg:confident aggregator}). 
Both consensus estimation and vote aggregation (answering the query) are noised with $\mathcal{N}(0, \sigma_1^2)$ and $\mathcal{N}(0, \sigma_2^2)$, respectively; where $\sigma_1, \sigma_2$ are tuned for better student accuracy.

\paragraph{DP-SGD} The \dpsgd extends standard stochastic gradient descent (SGD) with two additional steps to implement privacy guarantees. 
First, the individual data points' gradients are clipped to a maximum gradient norm bound $C$.
This bounds the gradients' sensitivity, which ensures that no data points can incur changes to the model above magnitude $C$.
After clipping, Gaussian noise with scale $\mathcal{N}(0, \sigma^2C^2)$ is added to mini-batches of clipped gradients. The noise distribution has zero mean and standard deviation proportional to a pre-defined noise multiplier $\sigma$ and the clipping norm $C$.
We detail the \dpsgd algorithm in \Cref{alg:dpsgd} in \Cref{app:dpsgd}.

To yield tighter privacy bounds, DP-SGD implements a privacy amplification through subsampling~\cite{beimel2010bounds}:
Training data points are sampled into mini-batches with a Poisson sampling per training iteration, in contrast to grouping the entire training data into mini-batches prior to every epoch as done in standard SGD.
Hence, the traditional concept of an epoch (as a full training on the entire training data) does not exist in DP-SGD.
Instead, each data point is sampled in every iteration according to a given sampling probability.
Privacy amplification through subsampling allows to scale down the noise $\sigma$ by the factor $L/N$ (with $L$ being the expected mini-batch size, $N$ the total number of data points, and $L \ll N$) while still ensuring the same $\varepsilon$ as with $\sigma$~\cite{kairouz2021practical} which is crucial to the practical performance (privacy-utility trade-offs) of DP-SGD.

\subsection{Pareto Efficiency}
\label{sec:background-pareto}
Let $\Theta$ be the set of all feasible baseline ML models with an element $\model \in \Theta$. A feasible model is one that is achievable through learning (optimization) over a given dataset. Consider $I$ to be the set of measurable trustworthiness objectives
such that the \textit{loss value}
of objective $i \in I$ is described by $\ell_i(\model)$. For instance, without loss of generality, $u_\text{priv} = \varepsilon$ where $\varepsilon$ is the achieved differential privacy budget from~\Cref{df:Differential Privacy}, and $l_\text{fair} = \hat{\Gamma}{(z, k)}$ is the demographic parity loss in~\Cref{sec:background-fairness}. We assume lower loss values are desirable.

\begin{definition}[Pareto Efficiency]   
\label{def:pareto}
$\model \in \Theta$ is \pe if there exists no $\model^{\prime} \in \Theta$ such that (a) $\forall i \in I$ we have $\ell_i\left(\model^{\prime}\right) \leq \ell_i(\model)$, and that (b) for at least one objective $j \in I$ the inequality is strict $\ell_j\left(\model^{\prime}\right) < \ell_j(\model)$.
\end{definition}

If an alternative model $\model'$ exists that satisfies conditions (a) and (b), we say $\model'$ \textit{Pareto dominates} $\model$. In this case $\model$ is \textit{\pie}, and therefore, is not on \textit{the \pf}. Intuitively, $\model$ is on the \pf, if we cannot improve objective $i$ without deteriorating another objective $j$.

\section{Resolving Tensions Between Fairness and Privacy}
\label{sec:tensions}

\begin{figure}[t]
\begin{minipage}[t]{0.2\textwidth}
		\begin{figure}[H]
        \includegraphics[width=\textwidth]{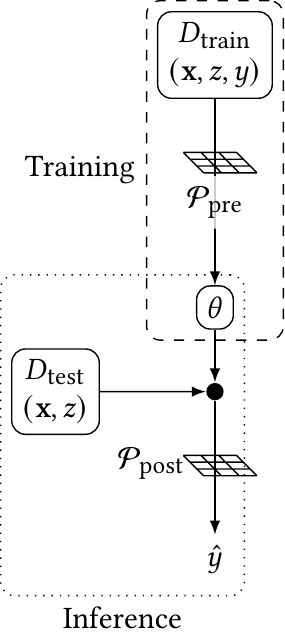}
        \end{figure}
\end{minipage}\hfill
\begin{minipage}[t]{0.38\textwidth}
    \begin{algorithm}[H] 
    \caption{\textbf{Pre-Processor $\mathcal{P}_\text{pre}$}}
    \label{alg:pre-processor}
    	\begin{algorithmic}[1]
             \setstretch{1.0}
    	   \Input{data point $x$, sensitive attribute $z$, true label $y$, subpopulation-class counts $m: \mathcal{Z} \times \mathcal{Y} \mapsto \mathbb{Z}_{\geq 0}$ } 
    	   \Require{minimum count $M$, fairness violation margin $\rfair$} 
    		\If{$\sum_{\tilde{y}} m({z, \tilde{y}}) < M$} \label{line:cold_start-pre}
        		\State $m(z, y) \gets m(z, y) + 1$
            	\State\Return $x$ 
            \Else
                \If{$\left(\frac{m(z, \hat{y})+1}{\left(\sum_{\tilde{y}} m(z, {\tilde{y}})\right)+1} - \frac{\sum_{\tilde{z}\neq z} m(\tilde{z}, \hat{y})}{\sum_{\tilde{z}\neq z, \tilde{y}} m(\tilde{z}, {\tilde{y}})}\right) < \rfair$}
                    \label{line:fairness_reject-pre}
                	\State $m(z, y) \gets m(z, y) + 1$
                	\State \Return $x$
                \Else
                	\State\Return $\bot$ 
                \EndIf
            \EndIf
    	\end{algorithmic}
    \end{algorithm}
\end{minipage}\hfill
\begin{minipage}[t]{0.38\textwidth}
    \begin{algorithm}[H] 
    \caption{\textbf{\textcolor{blue}{Post}-Processor $\mathcal{P}_\text{post}$}}
    \label{alg:post-processor}
    	\begin{algorithmic}[1]
             \setstretch{1.0}
    	   \Input{data point $x$, sensitive attribute $z$,\textcolor{blue}{predicted label $\hat{y}$}, subpopulation-class counts $m: \mathcal{Z} \times \mathcal{Y} \mapsto \mathbb{Z}_{\geq 0}$ } 
    	   \Require{minimum count $M$, fairness violation margin $\rfair$} 
    		\If{$\sum_{\tilde{y}} m({z, \tilde{y}}) < M$} \label{line:cold_start-post}
        		\State $m(z, y) \gets \textcolor{blue}{ m(z, \hat{y})} + 1$
            	\State \textcolor{blue}{\Return $\hat{y}$} 
            \Else
                \If{$\left(\frac{m(z, \hat{y})+1}{\left(\sum_{\tilde{y}} m(z, {\tilde{y}})\right)+1} - \frac{\sum_{\tilde{z}\neq z} m(\tilde{z}, \hat{y})}{\sum_{\tilde{z}\neq z, \tilde{y}} m(\tilde{z}, {\tilde{y}})}\right) < \rfair$}
                    \label{line:fairness_reject-post}
                	\State $m(z, y) \gets \textcolor{blue}{ m(z, \hat{y})} + 1$
                	\State \textcolor{blue}{\Return $\hat{y}$}
                \Else
                	\State\Return $\bot$ 
                \EndIf
            \EndIf
    	\end{algorithmic}
    \end{algorithm}
\end{minipage}
\caption{
\textbf{Demographic parity pre- and post-processor.} We depict the placement of the fairness mitigation (left).
While the pre-processor $\mathcal{P}_\text{pre}$ (\Cref{alg:pre-processor}, middle) operates within training, the post-processor $\mathcal{P}_\text{post}$ (\Cref{alg:post-processor}, right) is applied at inference time.
The subpopulation-class count $m$ refers to the number of data points per-class within each of the subpopulation groups.
It is used to empirically estimate the demographic disparity $\widehat{\Gamma}(z, k)$, $m: \mathcal{Z} \times \mathcal{K} \mapsto \mathbb{Z}_{\geq 0}$.
After a \textit{cold-start phase} (line 1-3) in both algorithms, we start rejecting queries for $x$ if we have to few samples from a given class.
}
\label{fig:pipeline}
\end{figure}

For an impartially-specified model,
we would like to consider fairness and privacy on equal footing. However, this is hard to achieve since the types of guarantees and their mitigations are inherently different. Differential privacy is an algorithmic and often data-independent guarantee. Once provided in a pipeline, thanks to its post-processing\footnote{We note that ``post-processing'' is an overloaded term. In algorithmic fairness literature, it means that the fairness mitigation happens after a model is trained while in differential privacy, it is a property which states that processing the output of a differentially private mechanism does not incur additional privacy leakage. Unless stated explicitly, we use post-processing in its algorithmic fairness sense.} property~\cite{dwork_algorithmic_2013},
it ensures privacy in the rest of the pipeline. The same cannot be said of fairness mitigations, as most are defined with respect to the model training process (see~\Cref{fig:pipeline}) and are often data-dependent.
In this section, we discuss the challenges of implementing fairness into private training through pre-processing. We present a post-processor as an ad-hoc solution concept which can replace pre-processing, or be used alongside it. Since studying in-processing fairness methods requires knowledge of the training procedure, we defer their study until~\Cref{sec:methods}.

\subsection{Shortcomings of Providing Fairness Through Pre-Processing in Private Learning}
\label{sec:dist-shift}

\paragraph{Fairness Degradation through Private Training}

Consider a model $\model$ with no fairness mitigations. Assume we use a demographic parity pre-processor (such as \Cref{alg:pre-processor}) to achieve zero demographic parity violations ($\gamma=0$). If we wish to make the model $\model$ private, we need to introduce randomization through adding a calibrated amount of noise either to the model or its output~(see~\Cref{sec:background-privacy}). In either case, noising has the effect of flipping predicted labels, such that the demographic parity violations might increase. Put differently, differential privacy may cause a \textit{\ds} in the predictions, which our earlier fairness mitigation cannot account for.

\paragraph{Privacy Cost of Fairness Pre-Processing}
\label{sec:priv-processing}
Fairness pre-processing can lead to increased privacy costs during private training.   
A consequence of differential privacy is the privacy consumption regime~\cite{mironov_renyi_2017}: just by observing the data for the purposes of equalizing a fairness measure between subpopulations, we may consume from the privacy budget.\footnote{Note that this disadvantage does not hold for fairness post-processing which does not incur additional privacy costs due to the differential privacy post-processing property.}
This budget could otherwise be spent, for instance, on more training passes on data to yield higher accuracy. 
We formalize this observation in~\Cref{thm:priv-pre-processing} for the case when a universal ordering exists. 
We defer the proof to \Cref{sec:preprocessing-proof}.
\begin{theorem}
\label{thm:priv-pre-processing}

Assume the training dataset $D = \{(\mathbf{x}, z, y) \mid \mathbf{x} \in \mathcal{X}, z \in \mathcal{Z}, y \in \mathcal{Y} \}$ is fed through the demographic parity pre-processor $\mathcal{P}_{pre}$ (as in \Cref{alg:pre-processor} without the minimum subgroup size constraint and applied offline retroactively) following an ordering defined over the input space $\mathcal{X}$. Let $\mathcal{P}_{pre}$ enforce a maximum violation $\gamma$, and $|Z| = 2$. Suppose now $\mathcal{M}$ is an $(\varepsilon,\delta)$ training mechanism, then $M \circ \mathcal{P}_{pre}$ is $(K_{\gamma} \varepsilon, K_{\gamma} e^{K_{\gamma} \varepsilon} \delta)$-DP where $K_\gamma = 2 + \left\lceil \frac{2\gamma}{1 - \gamma}\right\rceil$.

\end{theorem}

\subsection{Improving Fairness-Privacy Interplay Through Additional Pre- and Post-Processing}
\label{sec:solutions_S1_S2}

\paragraph{\textbf{S1}: Fairness Pre-Processor}
\label{sec:semi-supervised}
On its own, \Cref{thm:priv-pre-processing} is a negative result: at least a subset of pre-processing mitigations will degrade the privacy guarantee. 
To avoid paying a factor of two (or more) on the privacy analysis, we propose extending the fairness pre-processor to perform an \textit{un-supervised fairness calibration via a public dataset}.
By leveraging the information from such a public dataset, one can ensure that demographic parity holds without paying the additional privacy costs.
Importantly, since estimating demographic parity does not require a ground-truth label, we can do so in an un-supervised fashion.
In this sense, the public dataset serves as an independent calibration set.

\paragraph{\textbf{S2}: Fairness Post-Processor} 
Our solution for the \ds is to rely on a post-processor.
We know, through the differential privacy post-processing property, that this type of mitigation does not cost us any additional privacy budget. 
More importantly, since the mitigation occurs at the last stage of the ML pipeline (see~\Cref{fig:pipeline}); the fairness guarantee is ensured at test (inference) time in spite of the aforementioned label shift induced by differential privacy.

Our post-processor in~\Cref{alg:post-processor} implements a filtering mechanism that determines which queries to answer given a current estimates of the demographic parity $\Gamma(z, k)$. 
A query $x$ is rejected if answering it violates the maximum demographic parity violations $\gamma$.
The algorithm has a cold-start phase where the model releases decisions on all queries until at least $M$ queries for each subgroup $z$ have been answered.

\section{In Search of Impartial Algorithms}
\label{sec:methods}
In designing a trustworthy system, ideally, we want to avoid inefficiencies in every objective. 
The prior section highlights that certain ways of composing ML pipelines are more efficient than others: while privacy degrades through fairness pre-processing it is not affected by fairness post-processing. 
This section focuses on in-processing mechanisms, which we left out in the prior section due to their dependence on the training framework. 
We start our study with DP-SGD---currently the predominant mechanism for privacy-aware machine learning. 
\subsection{Integrating Fairness into DP-SGD}
\label{sec:dpsgd}

Having presented the necessary background in DP-SGD in~\Cref{sec:background-privacy}, here, we first propose how one can incorporate a demographic parity fairness in-processor within DP-SGD.
Given that DP-SGD builds on optimizing a learning loss in an unconstrained optimization setup, a valid solution consists in incorporating a fairness regularizer~\cite{kamishima_fairness-aware_2011} within the learning loss. 
Such a fairness regularizer always needs to estimate a (group) fairness statistic---in our case, the demographic disparity---on the training data. 
In DP-SGD, this is challenging for two main reasons:
1) Since, in SGD training is conducted in mini-batches, we need to ensure that the mini-batches are representative of the whole dataset and large enough to produce an accurate estimation of the fairness statistic.
2) Using the training data for additional fairness estimation during training consumes additional privacy budget which could otherwise be spent training for more iterations on the data, and thereby, improving model accuracy.

On a first glance, 1) might be easily resolved, for example, by stratified sampling~\cite{stratified} of mini-batches, and using larger mini-batch sizes (or switching to full-batch gradient descent); respectively. 
However, from a privacy perspective, these changes to the sampling procedure will severely degrade the improved privacy-utility trade-offs from the privacy amplification through subsampling in DP-SGD (see \Cref{sec:background-privacy}). 
Instead, we propose addressing 1) through regularization during training and 2) by assessing demographic parity on an unlabeled public dataset, as sketched in \textbf{S1} in the prior section. 
The resulting changes yield our novel \fdpsgd algorithm.

\subsection{Our \fdpsgd}
\label{sec:fdpsgd}

We introduce \fdpsgd (\Cref{alg:fdpsgd}), which extends \dpsgd with a demographic parity fairness regularizer (DPL). 
In order to avoid paying an extra privacy cost for inferring the fairness measure over the training set, we do so over a public dataset. 
The resulting demographic parity loss is: 
\begin{align} \label{eq:DPL}
    \operatorname{DPL}(\model; X_\text{public}) = \max_{k}\max_{z}\widehat{\Gamma}(z, k) = \max_{k}\max_{z} \left\{\frac{|\{\hat{Y}=k, Z = z \}|}{|\{Z = z\}|} - \frac{|\{\hat{Y}=k, Z \neq z\}|}{|\{Z \neq z\}|} \right\}
\end{align}
where $\widehat{Y} = \model(X_\text{public})$ is the prediction of the private model $\model$ on the features $X_\text{public}$ of the public dataset $D_\text{pub}$. 

We need to preserve useful gradients when implementing the DPL for it to be effective. 
Calculating the DPL requires calculating the fairness violation, $\Gamma(z, k)$, which uses the predicted label $k$. 
An $\argmax$ is typically applied to the output to obtain the prediction, but the operation is not differentiable, and hence unsuited to obtain useful gradients. 
To overcome this limitation, we use a tempered softmax: $\operatorname{softmax}_{T}(x_{i}) = \frac{\exp{x_{i}/T}}{\sum_{\tilde{i}} \exp{x_{i}/T}}$, where $T$ is the temperature. We set $T$ to a very small value (\eg 0.01) to make the tempered softmax close to the $argmax$ while keeping the overall loss differentiable. 

During training, \fdpsgd adds the DPL to the original loss function and scales it by a regularization factor, $\lambda$ which balances between both loss functions. 
Note that in contrast to the fairness violation $\gamma$, $\lambda$ is a \textit{pre-specified} model parameter while $\gamma$ reports the fairness violation obtained by the final trained model. 
At inference time, \fdpsgd also uses the post processor~\Cref{alg:post-processor} to ensure that the results satisfy the specified fairness constraint. 
The privacy analysis of \fdpsgd follows that of \dpsgd with Poisson Sampling~\cite{abadi2016deep, zhu_poission_2019} with no additional privacy costs resulting from fairness assessment on the public data (see \textbf{S1} in \Cref{sec:priv-processing}).

\subsection{Integrating Fairness into PATE}
\label{sec:pate}
\begin{figure}
	\includegraphics[width=0.65\textwidth]{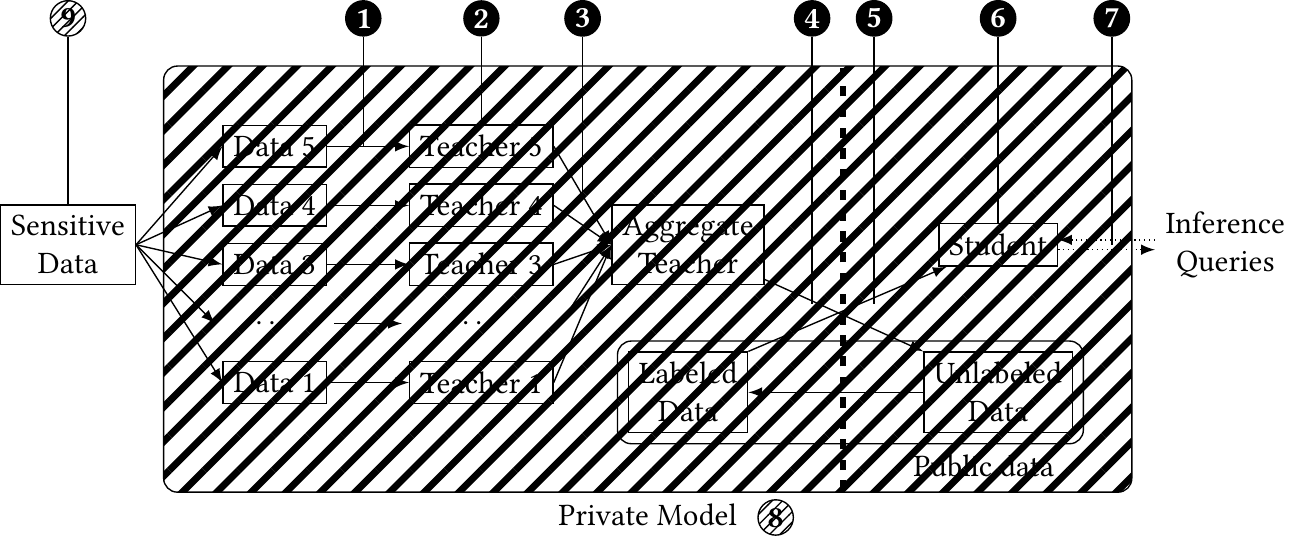}
    
    \caption{\textbf{Various ways to integrate fairness in PATE:}  For teachers: Pre-~\one/In-~\two/Post-Processing~\three. For the student: Pre-~\five/In-~\six/Post-Processing~\seven.
    A fair supervised privacy-preserving algorithm (\eg, \fdpsgd) replaces the private model (grey stripes) in-processing~\protect\eight, while a pre-processor applies to sensitive data directly~\protect\nine. Dashed line separates public and private data domains.
    Our \fpate's intervention occurs at \four.
}
    \label{fig:pate}
\end{figure}

Given that \fdpsgd has only a few degrees of freedom where fairness measures can be implemented, in the next section, we turn our study to the more complex PATE framework.
The modular design of PATE (see~\Cref{fig:pate}) allows us multiple points of fairness integration which we detail these in the following. These can potentially lead us to achieve stronger improvements to fairness-privacy tradeoff. 

\paragraph{Teacher-level~\one, \two, and~\three}
All three designs are non-impartial as they place fairness before privacy mitigation. 
Since in PATE, privacy is ensured at the level of aggregated teachers and not individual teachers, all the three alternatives can be seen as instances of the fairness pre-processor $\mathcal{P}_\text{pre}$ in \Cref{thm:priv-pre-processing} from the final student model's perspective. 
As a result, they all suffer from additional privacy leakage; on the level of teacher data~\one, model~\two, or vote~\three. 

\paragraph{Student-level~\five, \six, and~\seven}
These designs place privacy before fairness, therefore, they are also non-impartial. Thanks to differential privacy post-processing, the privacy budget remains unchanged. However, the drawbacks are in terms of fairness and accuracy. We discussed the former in \Cref{sec:dist-shift}. 
Regarding the impact on accuracy, remember that in the query phase of PATE, we incur a much smaller privacy cost for rejecting a query than for answering it (see~\Cref{sec:background-privacy}). 
Now consider the scenario in~\five~where a query is labeled but is ultimately rejected due to a fairness violation. In this case, the extra budget incurred for answering the query is wasted. Saving this budget could have allowed us to answer more queries, thus enabling higher student accuracy. Therefore,~\five~is not \pe.  We note that our demographic parity post-processor in~\Cref{alg:post-processor} is suitable for~\seven~but, on its own, still inefficient. We demonstrate the inefficiencies of~\six~and~\seven~empirically in~\Cref{sec:fairpate-design-validation}. 

\subsection{Our \fpate}
\label{sec:fairpate}
\begin{wrapfigure}{r}{0.45\textwidth}
    \vspace{-0.8cm}
    \begin{minipage}{0.45\textwidth}
    \begin{algorithm}[H]
    \setstretch{1.0}
    \caption{\textbf{-- Confident\&Fair-GNMax Aggregator} 
    }
            \label{alg:fair-pate}
        \begin{algorithmic}[1]
           \Input{query data point $x$, sensitive attribute $z$, predicted class label $k$, subpopulation subclass counts $m: \mathcal{Z} \times \mathcal{K} \mapsto \mathbb{Z}_{\geq 0}$} 
           \Require{minimum count $M$, threshold $T$, noise parameters $\sigma_1$, $\sigma_2$, fairness violation margin $\rfair$} 
        
            \If{$\max_j\{n_j(x)\} + \mathcal{N}(0, \sigma_1^2) \geq T$} 
                \State $k \gets \arg\max_j \left\{ n_j(x) + \mathcal{N}(0, \sigma_2^2)\right\}$ 
                \label{line:noisy_argmax_fairpate}
                \If{$\sum_{\tilde{k}} m({z, \tilde{k}}) < M$} 
          \label{line:cold-start}
                    \State $m(z, k) \gets m(z, k) + 1$
                    \State\Return $k$
                \Else
                        \If{$\left(\frac{m(z, k)+1}{\left(\sum_{\tilde{k}} m(z, {\tilde{k}})\right)+1} - \frac{\sum_{\tilde{z}\neq z} m(\tilde{z}, k)}{\sum_{\tilde{z}\neq z, \tilde{k}} m(\tilde{z}, {\tilde{k}})}\right) < \rfair$}
                        \label{line:fairness_reject}
                          \State $m(z, k) \gets m(z, k) + 1$
                          \State \Return $k$
                        \Else
                          \State\Return $\bot$ 
                        \EndIf
                    \EndIf
            \Else
                \State\Return $\bot$      
            \EndIf
        \end{algorithmic}
    \end{algorithm}
    \end{minipage}
    \end{wrapfigure}
Having discussed all of the alternatives, it is clear that the only viable for the fairness mitigation option is at~\four, namely,  at the level of aggregate teacher. This choice reflects the impartiality principle: the fairness mitigation occurs exactly at the point where PATE privacy-preserving mechanisms are implemented---avoiding the aforementioned inefficiencies.

\subsubsection{Confident\&Fair-GNMax}
Our new aggregation mechanism for \fpate, which we call \emph{Confident\&Fair-GNMax} (CF-GNMax) extends PATE's standard GNMax algorithm (\Cref{alg:confident aggregator}) with the idea of rejecting queries due to their privacy cost to also rejecting queries due to their disparate impact on fairness. 
Concretely, CF-GNMax, integrates an additional demographic parity constraint within the aggregator which allows rejecting queries on the basis of fairness, see 
\Cref{alg:fair-pate}.
The algorithm checks potential violations of demographic disparity violations and maintains an upper bound $\rfair$ on them in the course of answering PATE queries (\Cref{line:fairness_reject}). The goal is to bound $\Gamma(z, k)$ but which, in practice, needs to be empirically estimated.
Concretely, we measure demographic disparity
$\widehat{\Gamma}(z, k)$ using the counter $m: \mathcal{Z} \times \mathcal{K} \mapsto \mathbb{Z}_{\geq 0}$ which tracks per-class, per-subgroup decisions. 

Care must be taken to produce accurate $\Gamma(z, k)$ estimations: with few samples, $\widehat{\Gamma}(z, k)$ may be a poor estimator of $\Gamma(z, k)$. 
Therefore, we introduce a \textit{cold-start} stage where there are not yet enough samples to estimate $\widehat{\Gamma}(z)$ accurately. We avoid rejecting queries due to the fairness constraint at this stage.
Concretely, we require at least, on average, $M$ samples from the query's subgroup  before we reject a query on the basis of fairness (line~\ref{line:cold-start}).

\subsubsection{Student-Preprocessor}
\label{sec:pate-student-preprocessor}
\fpate can introduce a \ds in the training data of the student model due to rejecting queries to stay within the fairness constraint.
Thus, the student model needs to implement a similar procedure at inference time when answering queries. 
\fpate uses a post-processor as the one introduced in \Cref{alg:post-processor}, which mirrors the fairness constraint used to query the teachers while performing inference on the student.

\subsubsection{Privacy Analysis}
\label{sec:privacy-analysis}

\fpate's query phase (CF-GNMAX, \Cref{alg:fair-pate}) has two main differences with PATE's (C-GNMAX, \Cref{alg:confident aggregator}). 
First, it consists of two stages: a cold-start stage and a rejection/acceptance stage. 
Second, in \fpate, a query may not be answered not only because it is too privacy-costly, but also because answering it would violate the fairness ($\gamma$-DemParity) constraint. 
During the cold-start stage, \fpate follows the same analysis as PATE's (\Cref{sec:pate-priv}).
When the cold start phase is finished (line ~\ref{line:cold-start}), we adjust the privacy analysis to account for the fact that in \fpate, a query $q_i$ can be additionally rejected if answering $q_i$ violates the maximum disparity gap. We can calculate \fpate's probability of answering query $q_i$ as:

\begin{align}
        \mathbb{P}[\text{answering } q_i(z, k)] &= \begin{cases}
            0  &\frac{m(z, k)+1}{\left(\sum_{\tilde{k}} m(z, {\tilde{k}})\right)+1} - \frac{\sum_{\tilde{z}\neq z} m(\tilde{z}, k)}{\sum_{\tilde{z}\neq z, \tilde{k}} m(\tilde{z}, {\tilde{k}})} > \rfair \\
            \tilde{q} & \text{otherwise}
        \end{cases}
\end{align}
where $k$ is the noisy argmax  (\Cref{line:noisy_argmax_fairpate}), $\tilde{q}$ is calculated using~\Cref{prop:pate} in \Cref{sec:pate-priv} (as before), and the left side of the condition is simply calculating the new tentative demographic disparity violation $\Gamma(z, k)$ if the query is accepted. 

Note that, since the value of the counter $m(z, k)$ is only conditioned on the value of the noisy argmax, by the post-processing property of DP~\cite{dwork_dp}, $m(z, k)$ and by extension, \Cref{line:fairness_reject} do not add any additional privacy cost. In other words, rejecting queries on the basis of fairness, 
does not incur additional privacy cost.

\section{Empirical Evaluation}
We evaluate \fpate and \fdpsgd on multiple datasets and derive the \pfs between fairness, privacy, and accuracy. 
Based on the \pfs, we answer the following research questions (\textbf{RQ}s):
\textbf{RQ1}: What is the impact of the post-processing strategy (see \textbf{S2} in~\Cref{sec:solutions_S1_S2}) that filters queries on trade-offs obtained by \fpate and \fdpsgd?
\textbf{RQ2}: Where are fairness mitigations best implemented within the private ML pipeline?
\textbf{RQ3}: How do \fpate and \fdpsgd differ in performance?
\textbf{RQ4}: Can baseline specification (see Section~\ref{sec:background}) by a public entity, e.g. a regulatory body, be done without having direct access to a the private data.

\textit{Experimental Setup.}
We evaluate five datasets, namely
ColorMNIST~\citep{arjovsky2019invariant}, CelebA~\citep{liu2015faceattributes}, FairFace~\citep{karkkainenfairface}, 
UTKFace\citep{zhifei2017cvpr}, and
CheXpert\citep{irvin2019chexpert}. 
We refer to Table~\ref{tab:datasets} in Appendix E for details on these datasets. 
For both \fpate and \fdpsgd, we train multiple models to study the trade-offs between model privacy, fairness, and accuracy. We train models with different privacy budget $\varepsilon$ and fairness specifications.
Reminder that in \fpate, we apply a fairness constraint $\gamma$. In~\fdpsgd, we have an unconstrained optimization problem and control the regularization factor $\lambda$. In all cases, we make a distinction between specified and achieved privacy budgets and fairness gaps. We report exclusively achieved values of $\gamma$ and $\varepsilon$, as well as model accuracy, and coverage on these models.

\subsection{Findings}
In this section, we derive \pfs for \fpate and \fdpsgd, as well as several relevant baselines for comparisons. For conciseness, we address our various methods using the visual language adopted in~\Cref{fig:pate}. For clarity, we distinguish our main contributions \fpate (\four+\seven) and \fdpsgd (\eight+\seven).

\begin{table*}
    \footnotesize
    \centering
    \begin{tabular}{llcccccccc}
        \toprule 
         \textbf{Setting}  & \textbf{Method} & \makecell{\textbf{$\varepsilon$-(\textdownarrow)}\\\textbf{Budget}}  & \makecell{\textbf{Max }(\textdownarrow)\\ \textbf{Disparity}} & \makecell{\textbf{Acc.} (\textuparrow)} & \makecell{\textbf{Cov.} (\textuparrow)}  & \makecell{\textbf{$\varepsilon$-(\textdownarrow)}\\\textbf{Budget}}  & \makecell{\textbf{Max }(\textdownarrow)\\ \textbf{Disparity}} & \makecell{\textbf{Acc.} (\textuparrow)} & \makecell{\textbf{Cov.} (\textuparrow)} \\
         \midrule
         &&\multicolumn{4}{c}{\textbf{ColorMNIST}} & \multicolumn{4}{c}{\textbf{UTKFace}}\\
        \cmidrule(lr){3-6} \cmidrule(lr){7-10}
        \multirow{5}{*}{Fair} 
        &\fpate & 2.88 & 0.01 & \textbf{85.6} & 0.62 & 8.65    & 0.01     & \textbf{83.8} & 0.78 \\
        &\fdpsgd &\textbf{1.0} & 0.01 & 85.4 & 0.64 & \textbf{8.0} & 0.01 & 81.2 & 0.72 \\
        &\five~ + \seven &2.88 & 0.01 & 80.9 & \textbf{0.69} & 10.0     & 0.01     & 82.6  & \textbf{0.82}\\
        &\six~ + \seven  &2.88 & 0.01 & 84.6 & 0.63 & 10.0   & 0.01     & 81.4   & 0.74 \\
        \midrule
        \multirow{5}{*}{Private}
        &\fpate &1.0 & 0.10 & 73.8 & 1.00 & 2.0 & \textbf{0.13} & 74.0 & 0.98  \\
        &\fdpsgd &1.0 & 0.10 & \textbf{88.8} & 1.00 & 2.0 & 0.15 & \textbf{75.3} & 0.99 \\
        &\five~ + \seven &1.0 & 0.10 & 73.1 & 1.00 & 2.0 & 0.14 &  72.3 & 0.99 \\
        &\six~ + \seven &1.0 & 0.10 & 74.2 & 0.98 & 2.0 & 0.15 & 72.3 & 0.99 \\  
        \midrule 
        \multirow{5}{*}{Accurate}
        &\fpate &2.87 & 0.10 & 88.5 & 0.99 & 10.0 &0.2 &\textbf{82.9} &\textbf{0.97} \\
        &\fdpsgd &\textbf{2.0} & 0.10 & \textbf{88.6} & 0.99 & 10.0 & 0.1 & 81.3 & 0.96 \\
        &\five~ + \seven &2.88 & 0.10 & 88.1 & 1.0 & 10.0 & 0.01 & 82.6 & 0.82 \\
        &\six~ + \seven &3.0 & 0.10 & 88.5 & 0.99 & 10.0 & 0.15 & 81.6 & 0.92 \\
        
        \bottomrule
    \end{tabular}
    \caption{\textbf{Baseline Comparisons.} \five~ + \seven:  PATE with pre-processing. \six~ + \seven:  PATE with in-processing. \textdownarrow: the lower, the better; \textuparrow: the higher, the better. The values are chosen by setting a hard constraint on the variable that we want to optimize for for each objective (\textbf{Setting}). Within that constraint, we always optimize for accuracy first and the other variables second. We observe that \fpate and \fdpsgd achieve the highest accuracy in most settings.\five~ + \seven~ always achieves highest coverage when only optimizing for fairness.}
    \label{tab:baselines}
\end{table*}

\subsubsection{\textbf{RQ1: Post-processing is necessary for ensuring tight fairness gaps}} 
\begin{figure}
\centering
    \begin{subfigure}[t]{0.4\textwidth}
        \includegraphics[width=\textwidth,trim={0.6cm 3cm 4cm 3cm},clip]{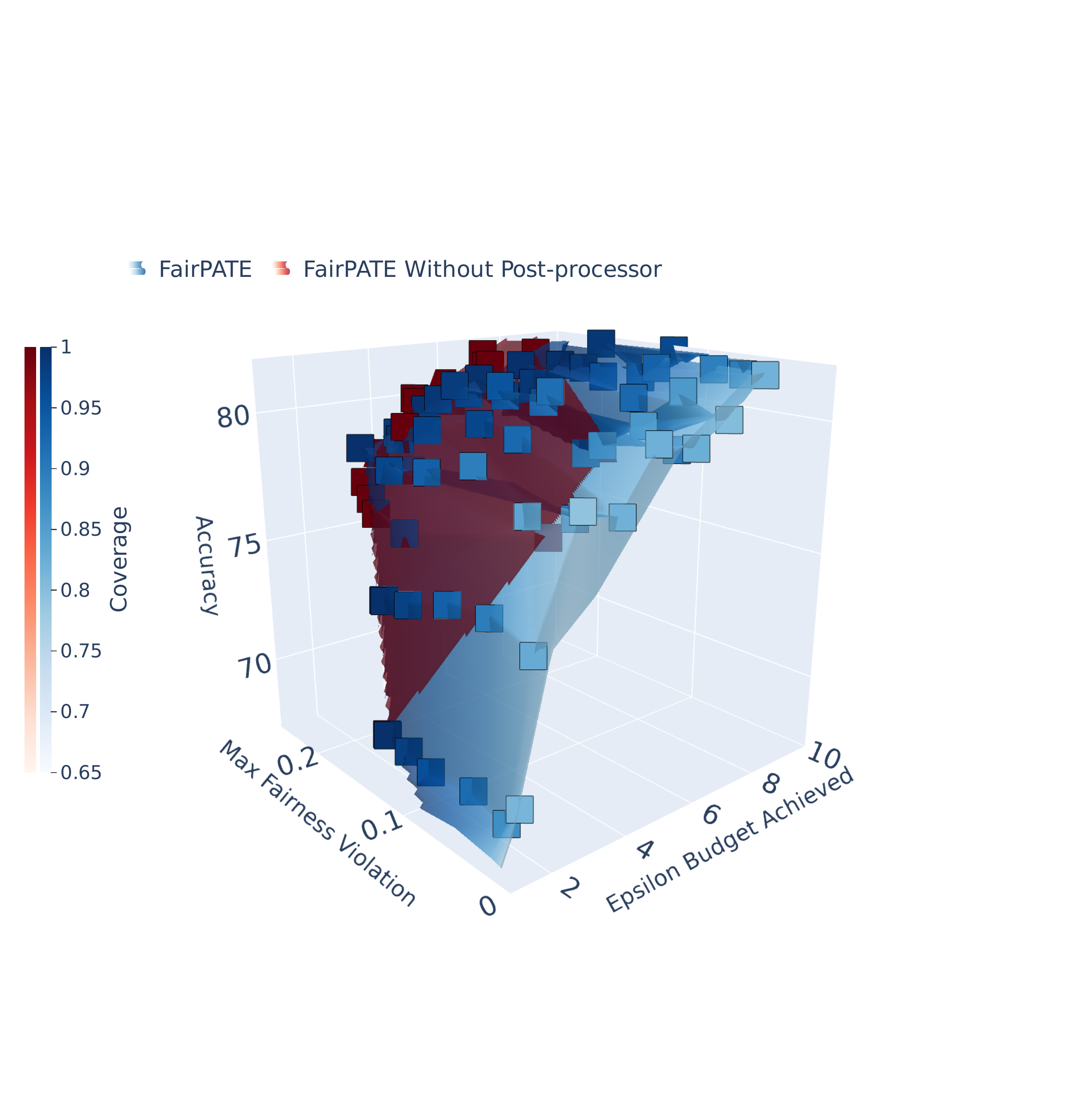}
        \caption{\fpate}
        \label{fig:fairPATE-vs-no-pre}
    \end{subfigure}
    \begin{subfigure}[t]{0.4\textwidth}
        \includegraphics[width=\textwidth]{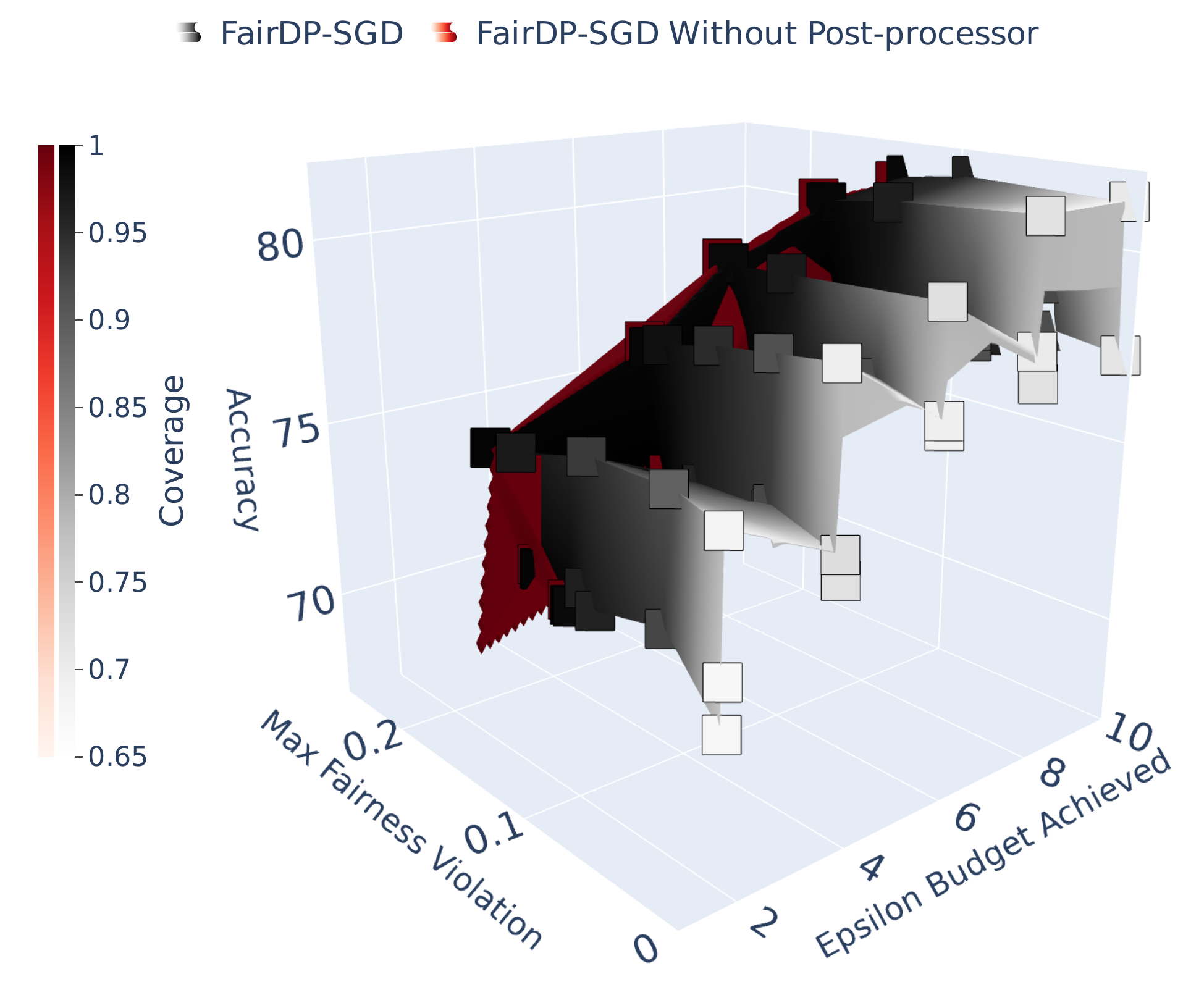}
        \caption{\fdpsgd %
        }
        \label{fig:fairDP-SGD-vs-no-pre}
    \end{subfigure}
\caption{\textbf{\fpate and \fdpsgd with vs. without post-processor on UTKFace.} Post-processor helps satisfy small fairness constraints while preserving model accuracy at the cost of answering fewer queries. 
}
\end{figure}

To study the necessity of it, we train \fpate and \fdpsgd models without the post-processor~\seven.
\fpate results are shown in \Cref{fig:fairPATE-vs-no-pre}. 
Without the post-processor, results span a smaller range of fairness violations. 
This is expected as \fpate introduces a \ds in its training data that should be mirrored in the test data~\Cref{sec:pate-student-preprocessor}, as done by the post-processor. 
The post-processor, thus, ensures that tighter fairness gaps are feasible. 
We plot \fdpsgd results with and without post-processor in \Cref{fig:fairDP-SGD-vs-no-pre}.
We observe that using demographic parity loss (DPL; see~\Cref{eq:DPL}),
we can achieve smaller fairness violations but the model utility decreases accordingly. In order to reach very small fairness gaps, we lose all utility as the model becomes increasingly inaccurate. 
The post-processor can preserve utility while satisfying tight fairness constraints at the cost of answering slightly fewer queries.  

\subsubsection{\textbf{RQ2: \fpate Pareto dominates similar designs in most contexts}}
\label{sec:fairpate-design-validation}
\begin{figure}
\begin{subfigure}[t]{0.4\textwidth}
    \centering
    \includegraphics[width=\textwidth]{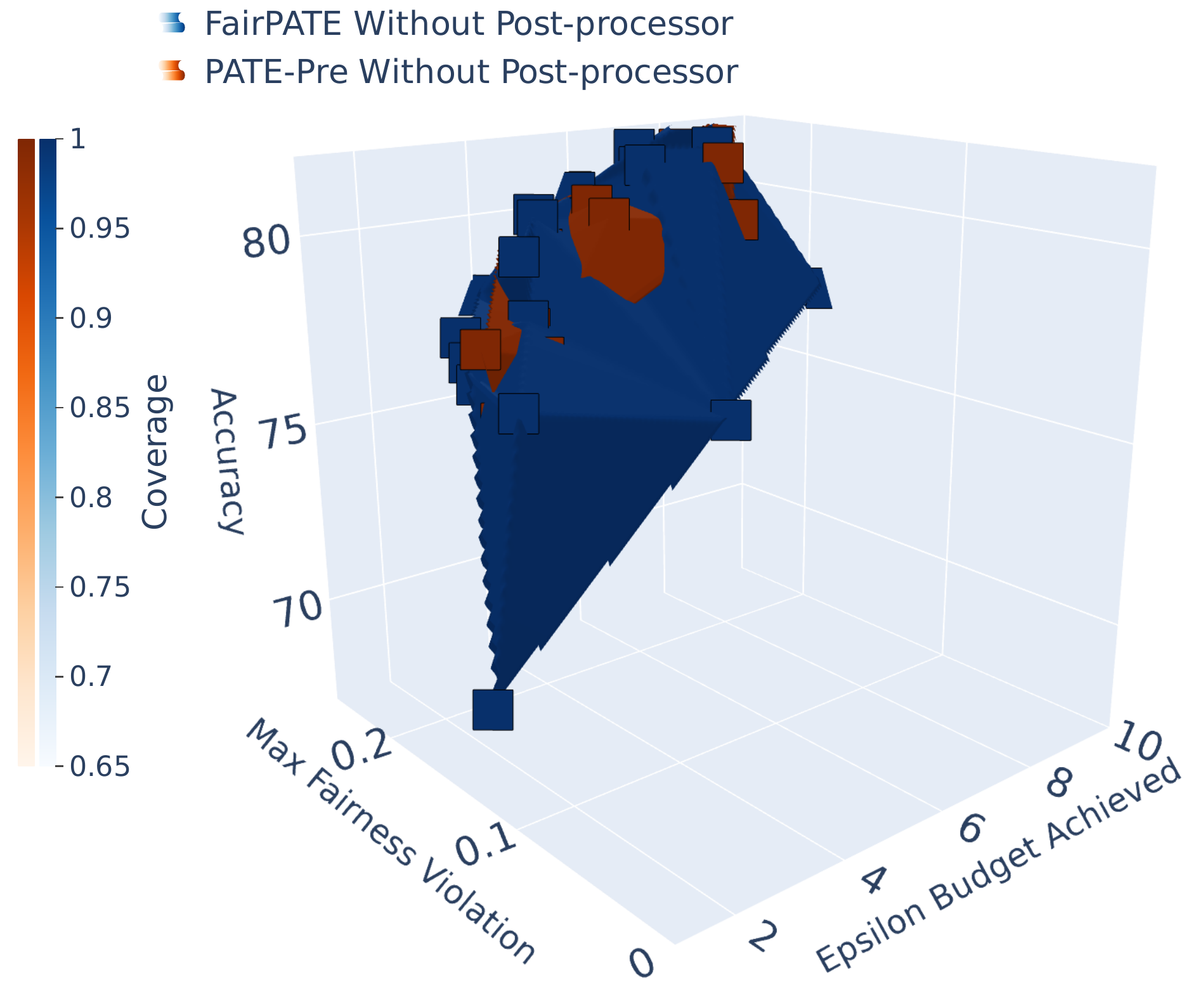}
    \caption{W/o post-processing: FairPATE~\four~ vs. pre-processed  PATE~\five.}
    \label{fig:fairPATE-vs-pre}
\end{subfigure}
\begin{subfigure}[t]{0.4\textwidth}
\centering
    \includegraphics[width=\textwidth]{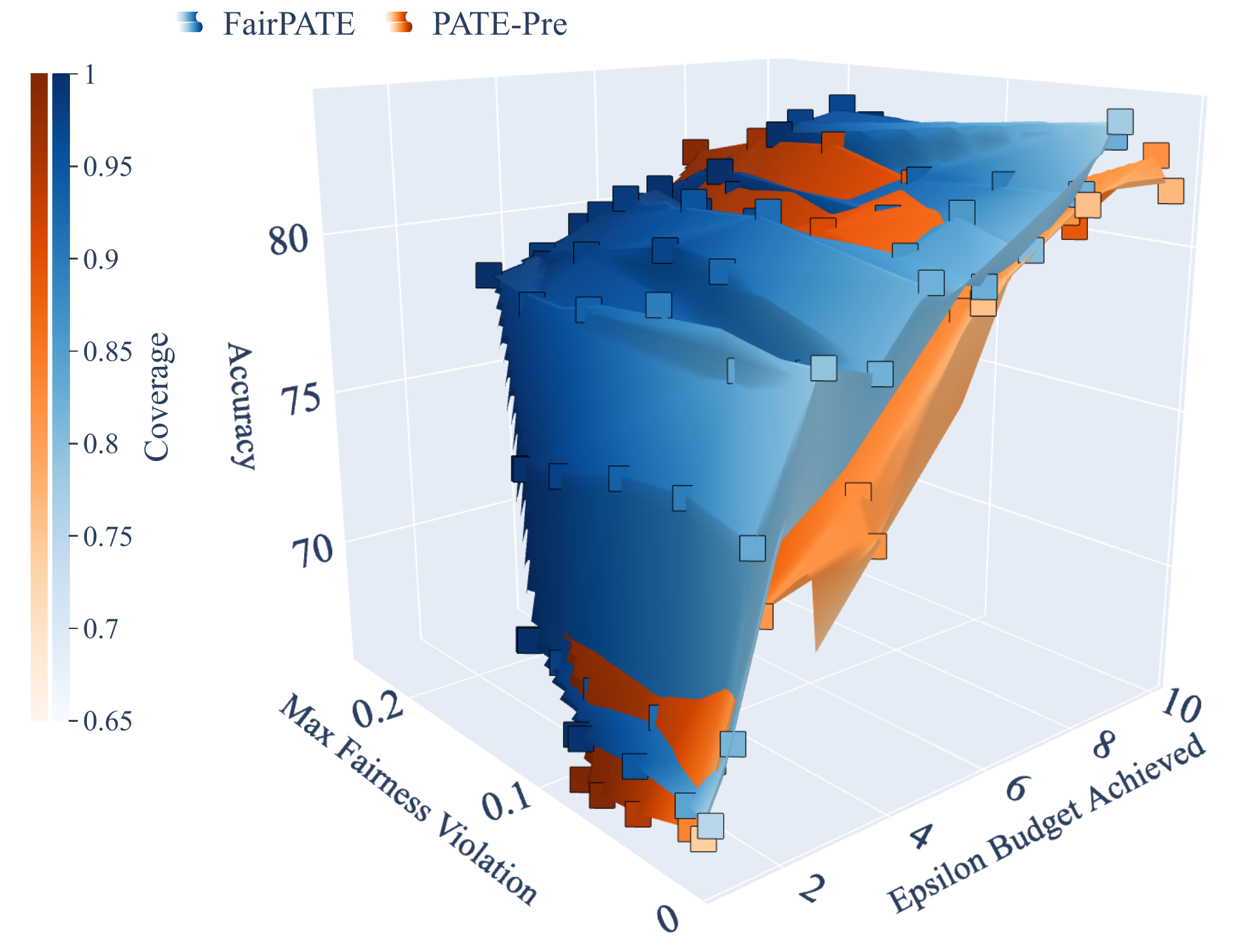}
    \caption{W/ post-processing: FairPATE~\four+\seven~ vs pre-processed  PATE~\five}
    \label{fig:fairPATE-vs-pre-in}
\end{subfigure}
\caption{\textbf{\fpate vs. pre-processed  PATE on UTKFace.} \fpate has higher accuracy than  PATE pre-processing in high privacy budget low fairness violation regions, and they have similar accuracy in other regions.
}
\end{figure}
\begin{figure}
\begin{subfigure}[t]{0.4\textwidth}
    \centering
    \includegraphics[width=\textwidth]{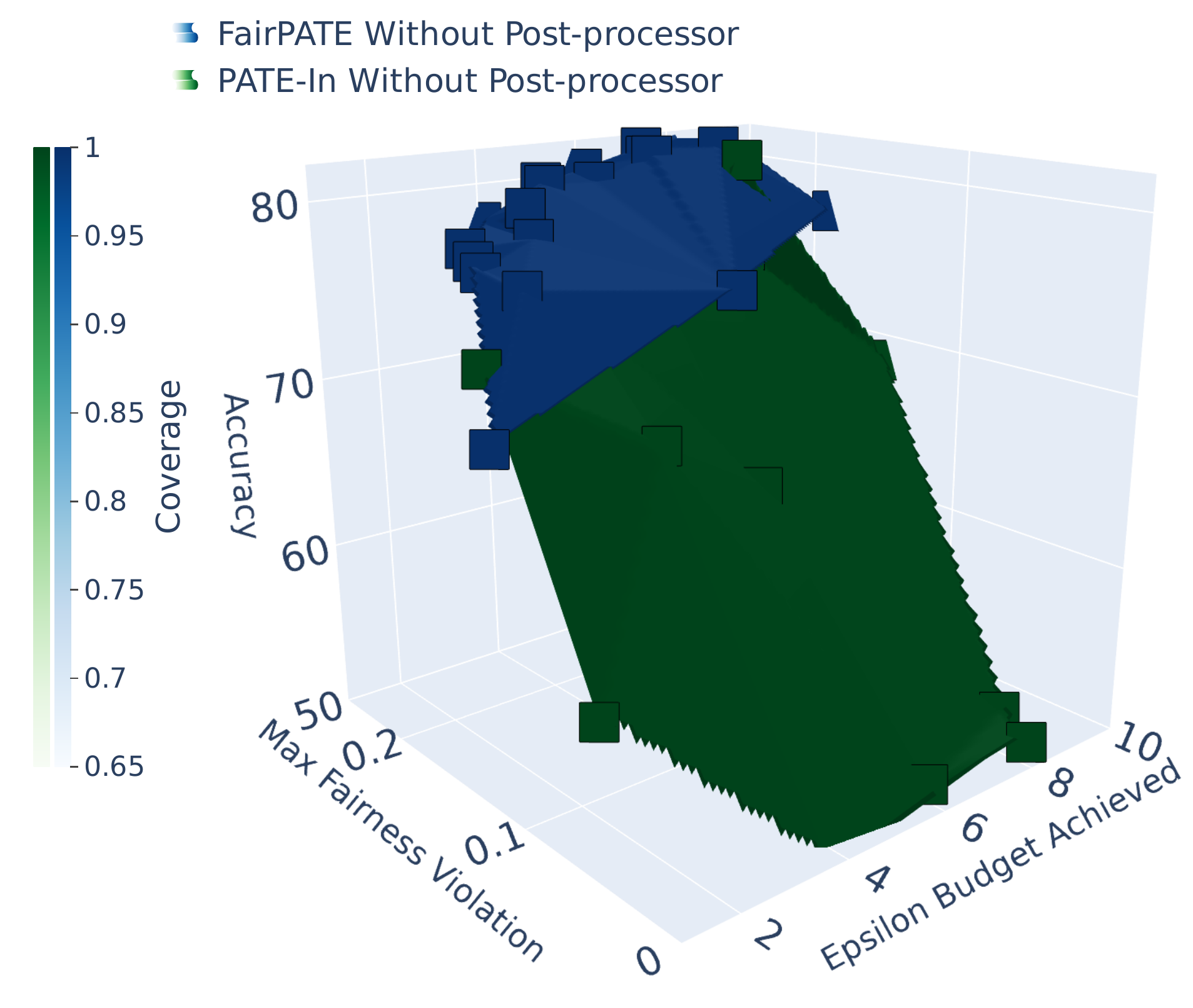}
    \caption{W/o post-processing: FairPATE~\four~vs. in-processed  PATE~\six}
    \label{fig:fairPATE-vs-in}
\end{subfigure}%
\begin{subfigure}[t]{0.4\textwidth}
\centering
    \includegraphics[width=\textwidth]{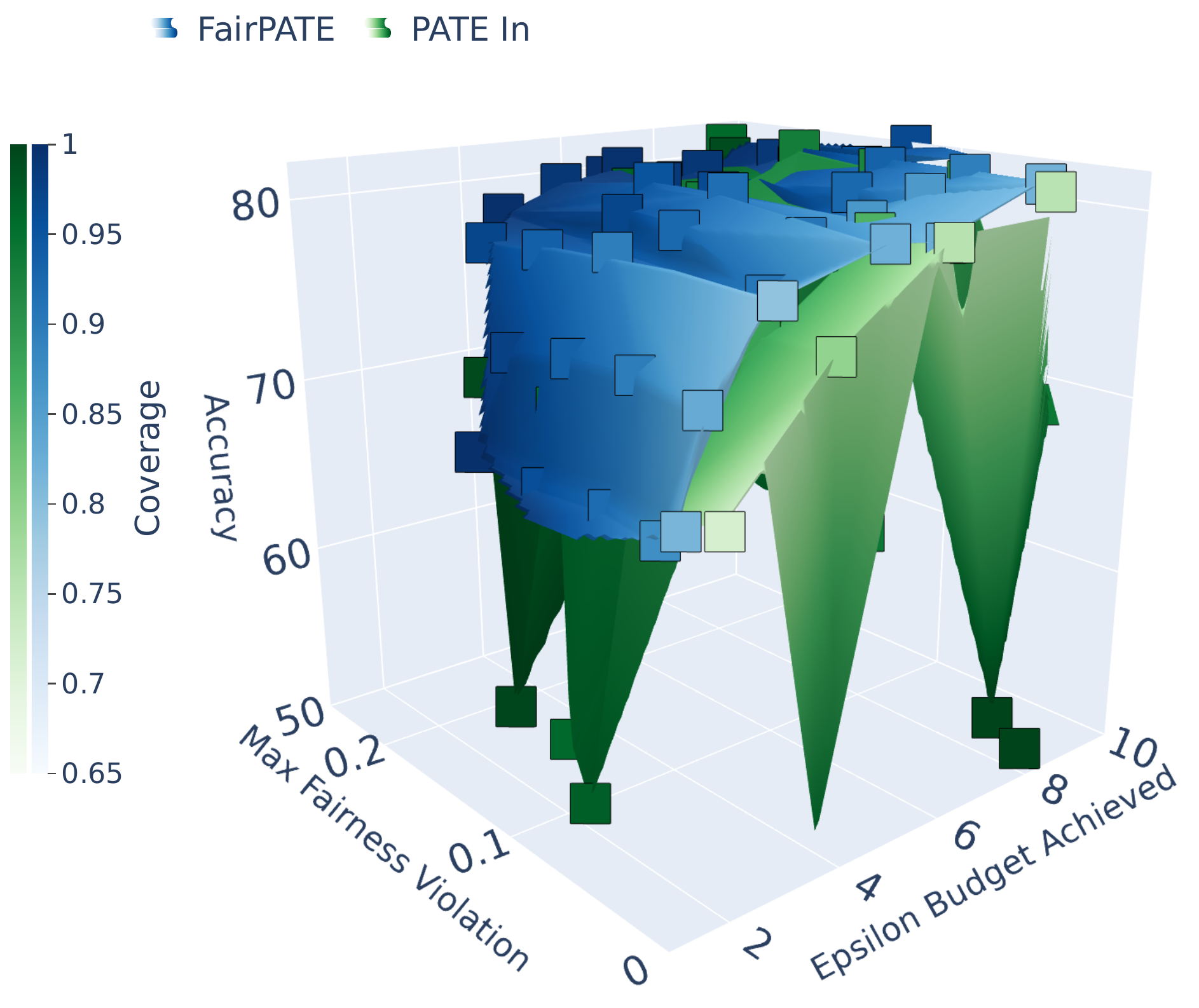}
    \caption{W/ post-processing: FairPATE~\four + \seven~vs. in-processed  PATE~\six + \seven}
    \label{fig:fairPATE-vs-in-in}
\end{subfigure}
\caption{\textbf{\fpate vs. in-processed  PATE on UTKFace.} \fpate has higher accuracy and coverage than  PATE in-processing in low fairness violation regions, and they have similar accuracy in other regions.
}
\end{figure}

The closest baseline to \fpate (\four+\seven) is querying with  PATE and then applying a pre-processor to the student model (\five + \seven). 
The important distinction between the two is that while selecting which queries to answer and which to reject, \four~ considers both fairness and privacy at the same time. In contrast, in \five,  PATE only enforces privacy while the fairness constraint is applied post-hoc. We first compare \four~ and \five~ without the post-processor. The respective student model results are shown in \Cref{fig:fairPATE-vs-pre}. 
We observe that the two methods achieve similar performances with higher privacy budget and higher fairness violation regions. 
This is because the fairness constraint is too loose to activate \fpate's fairness mechanism. 
In higher privacy budget and low fairness violation regions, \fpate yields student models of higher accuracy. 
This is because it is able to answer more queries since it saves privacy costs by rejecting queries due to fairness constraints. 
Then, we compare the two methods (\fpate (\four + \seven)~ and \five~ + \seven~) with the post-processor.
Results are shown in \Cref{fig:fairPATE-vs-pre-in}. 
In addition to the previous observations, we also note that the two methods perform similarly in very low privacy budget regions. 
In these regions, too few queries are answered to leave the cold-start phase and activate the actual fairness-based rejection mechanism.

We also compare \fpate to~\six, where we query with  PATE and employ an in-processing method in training the student model. 
For a controlled experiment, we use the same demographic parity loss as \fdpsgd for the fairness regularizer, namely DPL~\Cref{eq:DPL}.
The student model results in \Cref{fig:fairPATE-vs-in} highlight that models that use in-processing are able to achieve much smaller fairness violations without the post-processor.
Yet, utility decreases as fairness violation decreases.  With the post-processor (\Cref{fig:fairPATE-vs-in-in}), the two methods perform similarly in larger fairness violation regions, but \fpate perform better in smaller fairness violation regions with better accuracy and coverage.

\subsubsection{\textbf{RQ3: \fpate performs better than \fdpsgd, especially with higher privacy budgets}}
\begin{wrapfigure}[15]{r}{0.4\textwidth}
\vspace{-0.5cm}
        \includegraphics[width=0.4\textwidth]{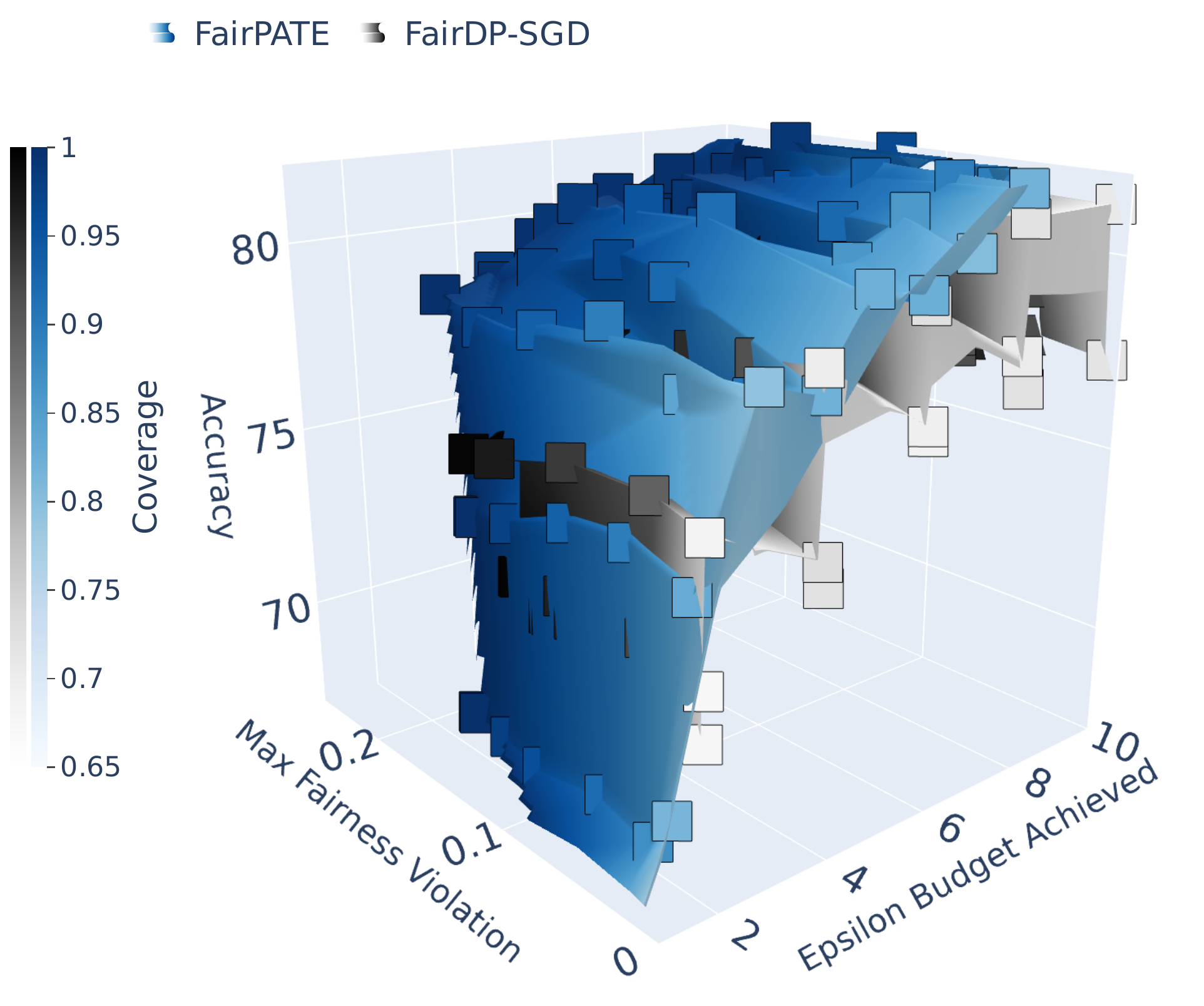}
        \caption{\textbf{\fpate vs \fdpsgd on UTKFace.} Our methods yield similar accuracy in low privacy budget or high privacy budget-high fairness violation regions. \fpate has higher accuracy and higher coverage in high privacy budget-low fairness violation regions.
        } %
    \label{fig:DPSGD-and-PATE}
\end{wrapfigure}
We compare our two methods, \fpate and \fdpsgd in \Cref{fig:DPSGD-and-PATE}. 
We observe that while they yield similar accuracy in low privacy budget regions, \fpate provides better accuracy in higher privacy budget regions. 
Additionally, in low fairness violation regions, \fpate obtaines higher accuracy and higher coverage.
This may be an artefact of regularization in \fdpsgd and could potentially be improved with hyper-tuning. 
However, this shows that it is easier to deriva a \pf for \fpate. We attribute this to the fact that performance (accuracy) of PATE (\fpate) is meaningfully correlated with the number of answered queries~(See~\Cref{sec:exp-answered-queries}). This allows for more fine-grained control on privacy costs, and thus a smoother \pf.

\begin{figure}
    \centering
    \begin{minipage}[t]{0.65\textwidth}
    \begin{figure}[H]
        \begin{subfigure}[t]{0.45\textwidth}
             \includegraphics[width=\textwidth]{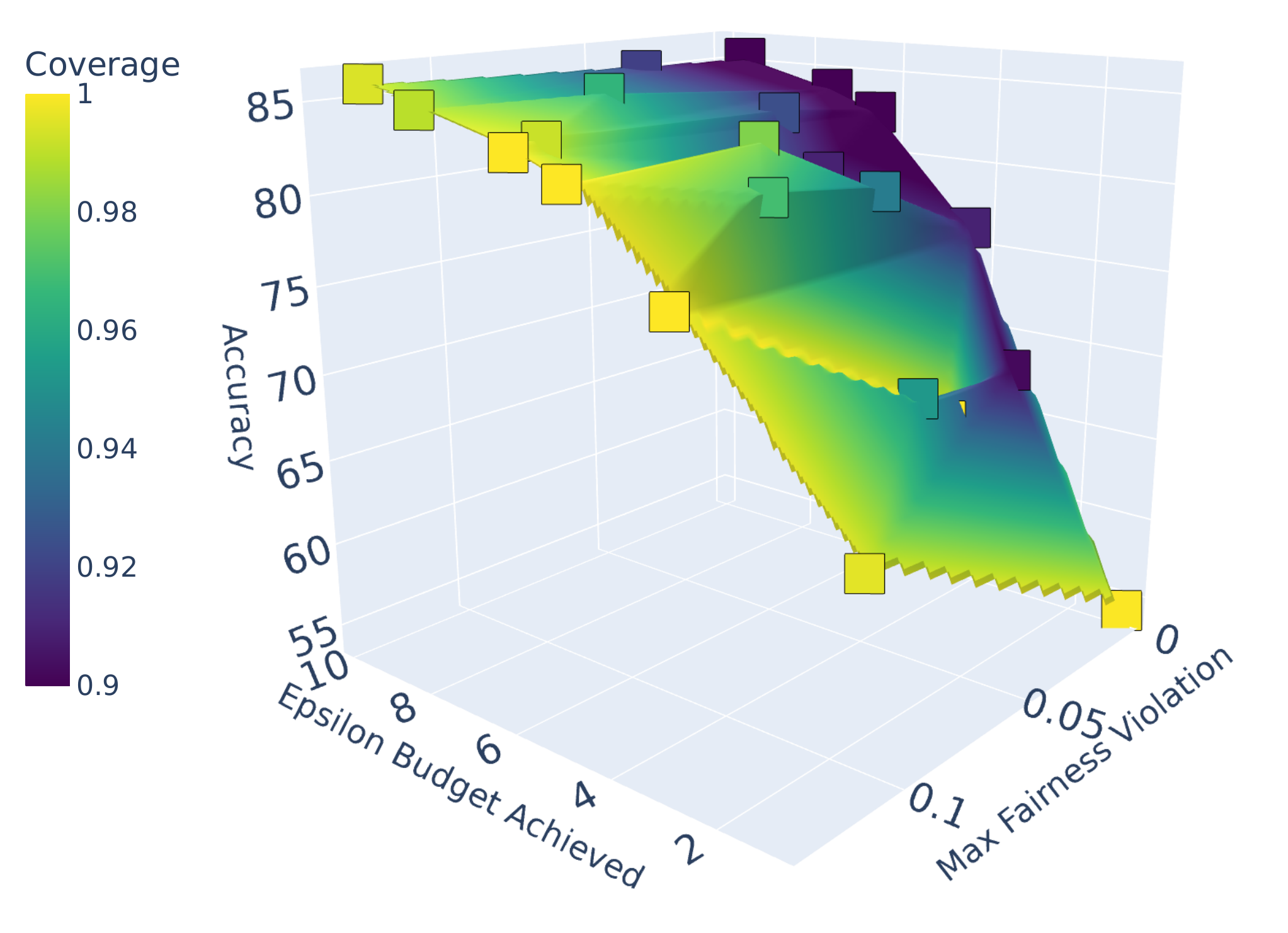}
             \caption{\fpate-CelebA}
              \label{fig:celeba-fairpate}
        \end{subfigure}
        \begin{subfigure}[t]{0.45\textwidth}
            \includegraphics[width=\textwidth]{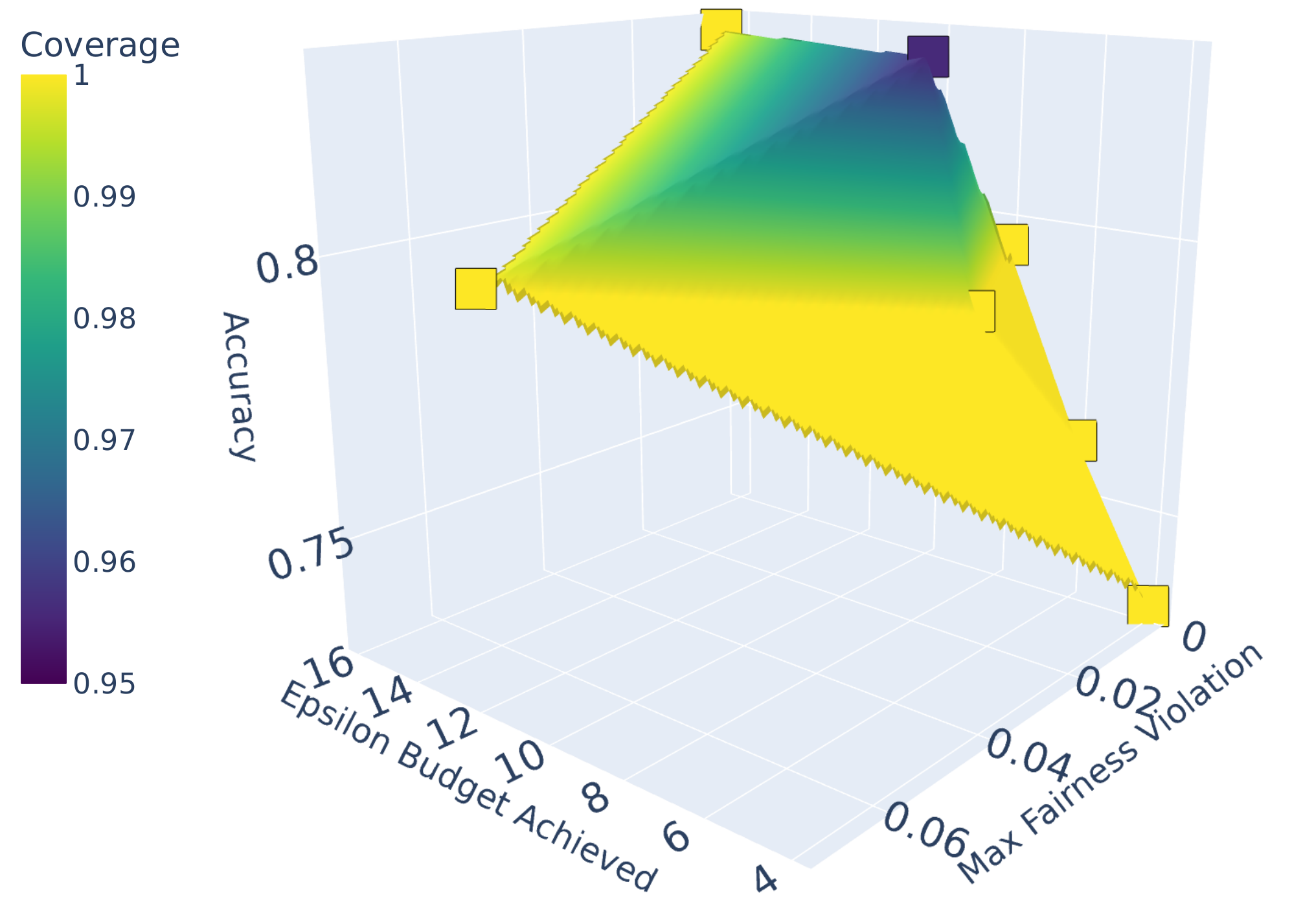}
            \caption{\fdpsgd-CheXpert}
            \label{fig:chexpert-fairpate}
        \end{subfigure}
        \caption{\textbf{\fpate on CelebA and CheXpert.} The figure plots the model results that are on the \pf. We observe that in both figures, model accuracy increases with higher privacy budget $\varepsilon$, and looser fairness constraints yield higher coverage.%
        }
    \end{figure}
    \end{minipage}\hfill
    \begin{minipage}[t]{0.32\textwidth}
    \begin{figure}[H]
    \includegraphics[width=\textwidth]{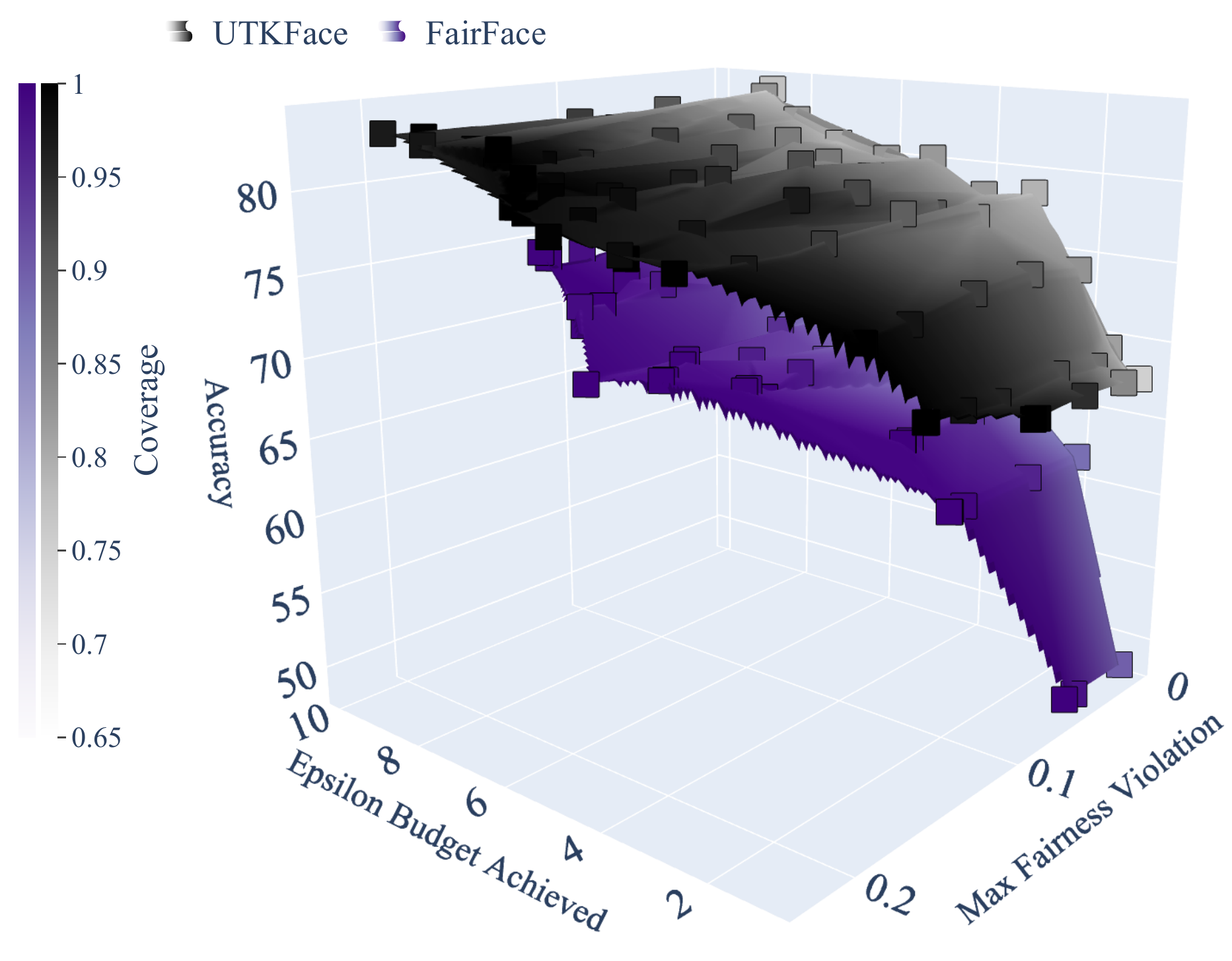}
    \caption{\textbf{Pareto Frontier of \fpate results on UTKFace and FairFace.} The two surface have very similar shapes despite the differences in accuracy.}
    \label{fig:fairface-utkface}
    \end{figure}
\end{minipage}
\end{figure}

\subsubsection{\textbf{RQ4: Baseline specification without direct data access is possible}}
We set out to explore whether the regulators can still make good decisions when specifying the baseline model parameters without access to the actual private dataset. Therefore, we compare the \pf surfaces obtained on different datasets.
\Cref{fig:celeba-fairpate} and \Cref{fig:chexpert-fairpate} plots the \pf surface from \fpate on CelebA and CheXpert respectively. \Cref{fig:fairface-utkface} plots the \pf from \fpate on UTKFace and FairFace.
Although the \pf surfaces show similar trends, the shapes are dataset dependent: different datasets show different trade-offs between the objectives.
However, we notice that the \pf surface shapes on UTKFace and FairFace are very similar. The classification task on both datasets is gender, with race as the sensitive attribute.  This shows that a regulator could use the \pf from a different dataset (which they have access to)  to design baseline specifications---as long as the dataset is sufficiently close in terms of classification task.

\section{Related Work}
\label{sec:related_work}

Due to the multiplicity of algorithmic fairness notions, as well as privacy; defining a benchmark to study fairness-privacy trade-offs is difficult. In this paper, we focus on discovering the \pf between demographic parity fairness (a \textit{group fairness} notion~\cite{fairMLBook}) and (central) differential privacy~\cite{dwork_algorithmic_2013}. While these objectives have a significant impact on each other (as we established in Section~\ref{sec:tensions}); each has been defined and developed independently of one~another.  

In contrast, there is a lineage of work that provides new definitions of fairness by characterizing the disparate impact of employing a privacy-aware mechanism~\cite{tran2021differentially, tran2021fairness}. 
While useful in their own regard, these new definitions do not alleviate the burden of satisfying established notions of fairness, such as demographic parity. 

Conceptually, the closest works to our setup are~\cite{jagielski2019differentially, mozannar2020fair} which assume different privacy notions. Both works strive to provide differential privacy (DP) with respect to the sensitive attribute. \citet{jagielski2019differentially} assumes a central notion of DP, while \citet{mozannar2020fair} assume a \textit{local} DP notion~\cite{bebensee_local_2019}. 
However, neither of the definitions used provide classical (approximate) differential privacy~\cite{dwork_algorithmic_2013} guarantees with respect to \textit{all features}. Furthermore, algorithms provided in these works, consider linear models and are optimized over tabular data. \fpate and \fdpsgd, on the other hand, are scalable deep-learning algorithms. Finally, neither prior work uses their formulations to derive a \pf.

\section{Limitations \& Conclusions}

Trustworthiness in machine learning is inherently a multi-objective endeavour. We acknowledge that as algorithm designers, we are only a part of the decision making process which likely occurs before any human judgement is passed. As such, it is imperative that (i) our design decisions should not limit (human) decision maker choices; and (ii) favour one objective over another. In this paper, we addressed the first challenge by providing a rich trade-off representation between the different objectives (fairness, privacy, and accuracy) in the form of a \pf. Our answer to the second challenge emerged as a design principle, which we called the impartiality principle.  We showed that models that break the impartiality principle are likely not on the \pf. 

Moving forward, the intuition behind our framework is pervasive to different formulations of what it means to be trustworthy. 
However, our current work assumes demographic parity as the fairness notion. We acknowledge that %
other group fairness and  individual fairness notions are prevalent in the literature. Adopting them would require %
adaptations of our algorithms. Similarly, we acknowledge the plethora of other privacy notions, including but not limited to, extensions or re-consideration of central (approximate) differential privacy. We leave their study to future work---having shown the value of ML mechanism design that supports obtaining Pareto frontiers.

\bibliographystyle{ACM-Reference-Format}
\bibliography{refs}

\printendnotes

\appendix
\section{Standard PATE Privacy Analysis}
\label{sec:pate-priv}
\citet{papernot2018scalable} use Rényi differential privacy (RDP)~\cite{mironov_renyi_2017} for accounting of the privacy budget expanded in answering each query. While the true privacy cost for each query is not known, an upporbound is estimated and summed over the course of the query phase. Answering queries stop when a pre-defined budget is exhausted. A student model is then trained on the answered queries.

\Cref{thm:rdp} establishes that the upperbound is a function of the  probability of \textit{not} answering a query $i$ with the plurality vote $i^*$. Unsurprisingly, this privacy cost function must tends to zero when the said event is very unlikely (\ie, strong consensus):

\begin{theorem}
\label{thm:rdp}
[From~\cite{papernot2018scalable}]
Let $\mathcal{M}$ be a randomized algorithm with $\left(\mu_1, \varepsilon_1\right)-R D P$ and $\left(\mu_2, \varepsilon_2\right)-RDP$ guarantees and suppose that given a dataset $D$, there exists a likely outcome $i^*$ such that $\operatorname{Pr}\left[\mathcal{M}(D) \neq i^*\right] \leq \tilde{q}$. Then the data-dependent Rényi differential privacy for $\mathcal{M}$ of order $\lambda \leq \mu_1, \mu_2$ at $D$ is bounded by a function of $\tilde{q}, \mu_1, \varepsilon_1, \mu_2, \varepsilon_2$, which approaches 0 as $\tilde{q} \rightarrow 0$.
\end{theorem}

In practice, \Cref{prop:pate} is used to find $\tilde{q}_i$ in \Cref{thm:rdp}, and $\mu_1, \mu_2$ are optimized to achieve the lowest upperbound on the privacy cost of each query for every order $\lambda$ of RDP.
\begin{prop}[From \cite{papernot2018scalable}]
\label{prop:pate}
For any $i^* \in[m]$, we have $
\operatorname{Pr}\left[\mathcal{M}_\sigma(D) \neq i^*\right] \leq \frac{1}{2} \sum_{i \neq i^*} \operatorname{erfc}\left(\frac{n_{i^*}-n_i}{2 \sigma}\right)
$, where \text{erfc} is the complementary error function.
\end{prop}

\section{Privacy Cost of Pre-Processing}
\label{sec:preprocessing-proof}

We provide the proof for \Cref{thm:priv-pre-processing}. 

\begin{proof}

We will proceed to show that using a pre-processing that sorts through data following some ordering defined over the whole input space
\footnote{An example of such ordering would be to order images based on their pixel values in some specified order of height, width and channel starting by checking the first pixel, then the second pixel, and so on.}
and, for any given label $y$, removes the last datapoints (following the ordering) in the majority subclass until it satisfies the $\gamma$-constraint will produce datasets at most $2 + K_\gamma = 2 + \left\lceil \frac{2\gamma}{1 - \gamma}\right\rceil$ apart. One then applies group privacy to obtain the final claim of the theorem.

Let $D' = D \cup {x^*}$, and the label of $x^*$ is $y^*$. We now proceed to analyze how far apart $\mathcal{P}_\text{pre}(D)$ and $\mathcal{P}_\text{pre}(D')$ can be. First note, they are the same on all labels not $y^*$, so we need only consider the difference on this label. First, let $m$ be the size of the minority subclass for label $y^*$ and let $m+c$ be the admissible size of the majority class. That is, we have $\frac{m}{2m +c} - \frac{m+c}{2m+c} < \gamma$. From this we can conclude $c = \lfloor \frac{\gamma}{1 - \gamma} 2m \rfloor$. Given this relation between the size of majority class a function of the minority class, we proceed to go through all logical cases to show the maximum difference is as claimed above.

Suppose $x^*$ belongs to the minority subclass for $y^*$ in $D$. Then we have $m \rightarrow m +1$ and hence $c \rightarrow \lfloor \frac{\gamma}{1 - \gamma} 2(m+1) \rfloor$. Thus we see $\mathcal{P}_\text{pre}(D')$ now admits one more point in the minority class of $y^*$ and at most $1 + \lceil \frac{2 \gamma}{1 - \gamma} \rceil$ more points to the the majority subclass (note we do not replace existing points as we follow the ordering on the input space). Thus the max change between $\mathcal{P}_\text{pre}(D)$ and $\mathcal{P}_\text{pre}(D')$ is $2 + \lceil \frac{2 \gamma}{1 - \gamma} \rceil$

Now suppose $x^*$ belongs to majority subclass for $y^*$ in $D$. In this case we have either $x^*$ appears early enough in the ordering that it now replaces another point in the majority class when applying $P$, or it is not added. In the former case, this mean we have changed $\mathcal{P}_\text{pre}(D)$ by $2$: we first removed a point and then added $x^*$. In the latter case, $x^*$ did not get added into the dataset, more so because of the ordering, $\mathcal{P}_\text{pre}(D') = \mathcal{P}_\text{pre}(D)$ as the order of points before $x^*$ is still the same. So in this case, once again, the change between $\mathcal{P}_\text{pre}(D)$ and $\mathcal{P}_\text{pre}(D')$ is less than $2 + \lceil \frac{2 \gamma}{1 - \gamma} \rceil$.

Thus we have by group privacy (see lemma 2.2 in \cite{vadhan2017complexity}) that $\mathcal{M} \circ \mathcal{P}_\text{pre}$ gives the claimed DP-guarantee, as we set $K_{\gamma} = 2 + \lceil \frac{2 \gamma}{1 - \gamma} \rceil$

\end{proof}

\section{Additional Background on Privacy-Preserving ML}
\label{app:dpsgd}
We include the standard \dpsgd algorithm (\Cref{alg:dpsgd}) and \fdpsgd (\Cref{alg:fdpsgd}) here for comparison. 

Details of the \fdpsgd algorithm is discussed in \Cref{sec:dpsgd}. 

\begin{algorithm}[h]
	\caption{Standard \dpsgd, adapted from~\cite{abadi2016deep}.}\label{alg:dpsgd}
	\begin{algorithmic}[1]
	\Require Private training set $D_\text{prv} = \{(x_i, y_i) \mid i \in [N_\text{prv}]\}$, loss function $ {\mathcal L}( \model, x_i)$,  Parameters: learning rate $\eta_t$, noise scale $\sigma$, group size $L$, gradient norm bound $C$. 
		\State {\bf Initialize} ${ \model}_0$ randomly
		\For{$t \in [T]$}
		\State {Sample mini-batch $L_t$ with sampling probability $L/N$} \Comment{Poisson sampling}
		\State {For each $i\in L_t$, compute ${\mathbf g}_t(x_i) \gets \nabla_{ \model_t} {\mathcal L}({ \model}_t, x_i) $}		\Comment {Compute gradient}
		\State {$\bar{{\mathbf g}}_t(x_i) \gets {\mathbf g}_t(x_i) / \max\left(1, \frac{\|{\mathbf g}_t(x_i)\|_2}{C}\right)$} \Comment{Clip gradient}
		\State {$\tilde{{\mathbf g}}_t \gets \frac{1}{|L_t|}\left( \sum_i \bar{{\mathbf g}}_t(x_i) + \mathcal{N}\left(0, \sigma^2 C^2 {\mathbf I}\right)\right)$} \Comment{Add noise}
		\State { ${ \model}_{t+1} \gets { \model}_{t} - \eta_t \tilde{{\mathbf g}}_t$} \Comment{Descent}
		\EndFor
		\State {\bf Output} ${ \model}_T$ and compute the overall privacy cost $(\eps, \delta)$ using a privacy accounting method.
	\end{algorithmic}
\end{algorithm}

\begin{algorithm}[h]
	\caption{\fdpsgd}\label{alg:fdpsgd}
	\begin{algorithmic}[1]
	\Require Private training set $D_\text{prv} = \{(x_i, y_i) \mid i \in [N_\text{prv}]\}$, Public calibration set $D_\text{pub} = \{ (\tilde{x}_i, z_i) \mid i \in [N_\text{pub}]\}$, loss function ${\mathcal L}({ \model})=\frac{1}{N}\sum_i {\mathcal L}( \model, x_i)$, Demographic Parity loss $\operatorname{DPL}(\model; D_\text{pub})$. Parameters: learning rate $\eta_t$, noise scale $\sigma$, group size $L$, gradient norm bound $C$. 
		\State {\bf Initialize} ${ \model}_0$ randomly
		\For{$t \in [T]$}
		\State {Sample mini-batch $L_t$ with sampling probability $L/N$} \Comment{Poisson sampling}
		\State {For each $i\in L_t$, compute ${\mathbf g}_t(x_i) \gets \nabla_{ \model_t} \left({\mathcal L}({ \model}_t, x_i) + \lambda\operatorname{DPL}(\model_t; D_{pub})\right)$}		\Comment {Compute gradient}
		\State {$\bar{{\mathbf g}}_t(x_i) \gets {\mathbf g}_t(x_i) / \max\left(1, \frac{\|{\mathbf g}_t(x_i)\|_2}{C}\right)$} \Comment{Clip gradient}
		\State {$\tilde{{\mathbf g}}_t \gets \frac{1}{|L_t|}\left( \sum_i \bar{{\mathbf g}}_t(x_i) + \mathcal{N}\left(0, \sigma^2 C^2 {\mathbf I}\right)\right)$} \Comment{Add noise}
		\State { ${ \model}_{t+1} \gets { \model}_{t} - \eta_t \tilde{{\mathbf g}}_t$} \Comment{Descent}
		\EndFor
		\State {\bf Output} ${ \model}_T$ and compute the overall privacy cost $(\eps, \delta)$ using a privacy accounting method.
	\end{algorithmic}
\end{algorithm}
\section{Fairness Metrics and Evaluations}
\label{sec:dem-parity-notions}
We evaluate and compare different ways to measure the demographic parity gap, $\Gamma(z, k)$. We then select one method to use in our implementations. 
We explore three different methods that compare the ratio between different sensitive groups to evaluate the chosen fairness metric.
\subsection{Demographic Parity Gap Measurements}
\begin{enumerate}
    \item Between Groups: This method computes and bounds the maximum difference between two pairs of sensitive groups.
    \begin{equation}
        \Gamma(z, k) := 
        max_{\tilde{z}} |\mathbb{P}[\hat{Y} = k | Z =  z] 
        - \mathbb{P}[\hat{Y} = k \mid Z = \tilde{z}]|.
    \end{equation}
    \item To Overall: This method computes and bounds the difference between each sensitive group and the total of all groups. 
    \begin{equation}
        \Gamma(z, k) := 
        \mathbb{P}[\hat{Y} = k | Z =  z] 
        - \mathbb{P}[\hat{Y} = k].
    \end{equation}
    \item To Overall Without Double Counting: This method computes and bounds the difference between each sensitive group and the total of all other groups. 
    \begin{equation}
        \Gamma(z, k) := 
        \mathbb{P}[\hat{Y} = k | Z =  z] 
        - \mathbb{P}[\hat{Y} = k \mid Z \neq z].
    \end{equation}
\end{enumerate}

To compare the three methods, we generate some synthetic data and run queries on them using each method to compare the results.

\subsection{Evaluation Results}

We first generate synthetic data with two classes and three sensitive groups. The distribution of the generated data is shown below. 

\begin{center}
\begin{tabular}{|l| c c c|} 
\hline
 Class/Sensitive Group & 0 & 1 & 2 \\ [0.5ex] 
 \hline\hline
 0 & 324 & 420 & 445 \\ 
 \hline
 1 & 287 & 274 & 250 \\ [1ex] 
 \hline
\end{tabular}
\end{center}

\subsubsection{By Group}
Total number of queries answered = 1661
\begin{center}
\begin{tabular}{|l| c c c|} 
\hline
 Class/Sensitive Group & 0 & 1 & 2 \\ [0.5ex] 
 \hline\hline
 0 & 315 & 364 & 361 \\ 
 \hline
 1 & 191 & 213 & 217 \\ [1ex] 
 \hline
\end{tabular}
\end{center}

\subsubsection{To Overall}
Total number of queries answered = 1832
\begin{center}
\begin{tabular}{|l| c c c|} 
\hline
 Class/Sensitive Group & 0 & 1 & 2 \\ [0.5ex] 
 \hline\hline
 0 & 324 & 395 & 361 \\ 
 \hline
 1 & 234 & 271 & 247 \\ [1ex] 
 \hline
\end{tabular}
\end{center}

\subsubsection{To Overall Without Double Counting}
Total number of queries answered = 1772
\begin{center}
\begin{tabular}{|l| c c c|} 
\hline
 Class/Sensitive Group & 0 & 1 & 2 \\ [0.5ex] 
 \hline\hline
 0 & 318 & 371 & 342 \\ 
 \hline
 1 & 218 & 244 & 229 \\ [1ex] 
 \hline
\end{tabular}
\end{center}

\subsection{Conclusion}
We decide to use the third method, to overall without double counting, as the comparison method. It is a balance between the by group method and the to overall method. We do not want the comparison method to be too strict, because then our algorithm would reject most queries due to fairness. On the other hand, we also do not want it to be too lenient that the fairness constraint is not enforced. One major drawback of the to overall method is that if most of the data is from one sensitive group, then that sensitive group would have too much influence over the overall class label distribution. 
\section{Experimental Setup}
\label{sec:setup}
We split each dataset into a training set, an unlabeled set, and a test set. 
The sizes of these three datasets are determined based on the dataset sizes specified in original PATE \cite{papernot2016semi, papernot2018scalable}, and adapted to the  difficulties of the prediction tasks. 
In \fpate, the training set is further split into equal partitions to train the teacher models. We train as many teachers as possible while still achieving good ensemble accuracy overall. In \fdpsgd, the whole training set is used to train the private model. %
The test set is used to evaluate the performance of the final model. %

For \fpate, we first train the teacher ensemble models, then query them with the public dataset, and aggregate their predictions using the \fpate algorithm. The student model is trained on the public dataset with obtained labels. The model architectures, as well as the parameters used in querying the teacher models are detailed in Table~\ref{tab:datasets} for each dataset, respectively. The model architectures are chosen by referencing what is used in related works for each dataset. 
\fdpsgd models are trained with the same model architecture as indicated in the table. 

\begin{table}[h]
\centering
\scalebox{0.8}{
\begin{tabular}{lccccccccccc}
\toprule
Dataset & Prediction Task & C & Sens. Attr. & SG & Total &  U & Model & Number of Teachers & $T$ & $\sigma_1$ & $\sigma_2$  \\
\midrule
    ColorMNIST~\citep{arjovsky2019invariant} & Digit & 10 & Color & 2 & \numprint{60000}& \numprint{1000} & Convolutional Network (\Cref{tab:CNN-ColorMNIST}) & 200 & 120 & 110 & 20\\
    CelebA~\citep{liu2015faceattributes} & Smiling & 2 & Gender & 2 & \numprint{202599} & \numprint{9000} & Convolutional Network (\Cref{tab:CNN-CelebA})& 150 & 130 & 110 & 10\\
    FairFace~\citep{karkkainenfairface} & Gender & 2& Race& 7& \numprint{97698} & \numprint{5000} & Pretrained ResNet50 & 50 & 30 & 30 & 10\\
    UTKFace~\citep{zhifei2017cvpr} & Gender & 2 & Race & 5 & \numprint{23705} & \numprint{1500} & Pretrained ResNet50 & 100 & 50 & 40 & 15\\
    CheXpert\citep{irvin2019chexpert} & Disease & 2 & Race& 3& \numprint{152847} & \numprint{4000} & Pretrained DenseNet121 & 50 & 30 & 20 & 10\\
    
\bottomrule
\end{tabular}
}
\caption{Datasets used for evaluation. Abbreviations: \textbf{C}: number of classes in the main task; \textbf{SG}: number of sensitive groups; \textbf{U}: number of unlabeled samples for the student training . Summary of parameters used in training and querying the teacher models for each dataset. The selection of $\sigma_1$ is in accordance with the threshold $T$. The selection process of $\sigma_2$, is shown in the Appendix~\ref{sec:setup}.%
}
\label{tab:datasets}
\end{table}

\begin{table}[h]
\begin{minipage}[t]{0.45\textwidth}
    \centering
    \begin{tabular}{lc}
    \toprule
    Layer & Description \\
    \midrule
    Conv2D with ReLU & (3, 20, 5, 1)  \\
    Max Pooling & (2, 2) \\
    Conv2D with ReLU & (20, 50, 5, 1) \\
    MaxPool & (2, 2) \\
    Fully Connected 1 & (4*4*50, 500) \\
    Fully Connected 2 & (500, 10) \\
    \bottomrule
    \end{tabular}
    \caption{Convolutional network architecture used in ColorMNIST experiments.}
    \label{tab:CNN-ColorMNIST}
\end{minipage} 
\begin{minipage}[t]{0.45\textwidth}
    \centering
    \begin{tabular}{lc}
    \toprule
    Layer & Description \\
    \midrule
    Conv2D & (3, 64, 3, 1)  \\
    Max Pooling & (2, 2) \\
    ReLUS & \\
    Conv2D & (64, 128, 3, 1)  \\
    Max Pooling & (2, 2) \\
    ReLUS & \\
    Conv2D & (128, 256, 3, 1)  \\
    Max Pooling & (2, 2) \\
    ReLUS & \\
    Conv2D & (256, 512, 3, 1)  \\
    Max Pooling & (2, 2) \\
    ReLUS & \\
    Fully Connected 1 & (14 * 14 * 512, 1024) \\
    Fully Connected 2 & (1024, 256) \\
    Fully Connected 2 & (256, 2) \\
    \bottomrule
    \end{tabular}
    \caption{Convolutional network architecture used in CelebA experiments.}
    \label{tab:CNN-CelebA}
\end{minipage}
\end{table}

We tune the amount of noise injected into the aggregation mechanism of \fpate by varying the standard deviation of the Gaussian distribution while ensuring the accuracy of the labels produced by the teacher ensemble models to maximize the accuracy of student models.

\section{The relationship between number of queries answered and student accuracy}
\label{sec:exp-answered-queries}
\begin{figure}[h]
\centering
    \begin{subfigure}[b]{0.4\textwidth}
         \includegraphics[width=\textwidth,clip]{utkface_fig2}
     \end{subfigure}
     \begin{subfigure}[b]{0.4\textwidth}
        \includegraphics[width=\textwidth,trim=2cm 2cm 3cm 3.5cm,clip]{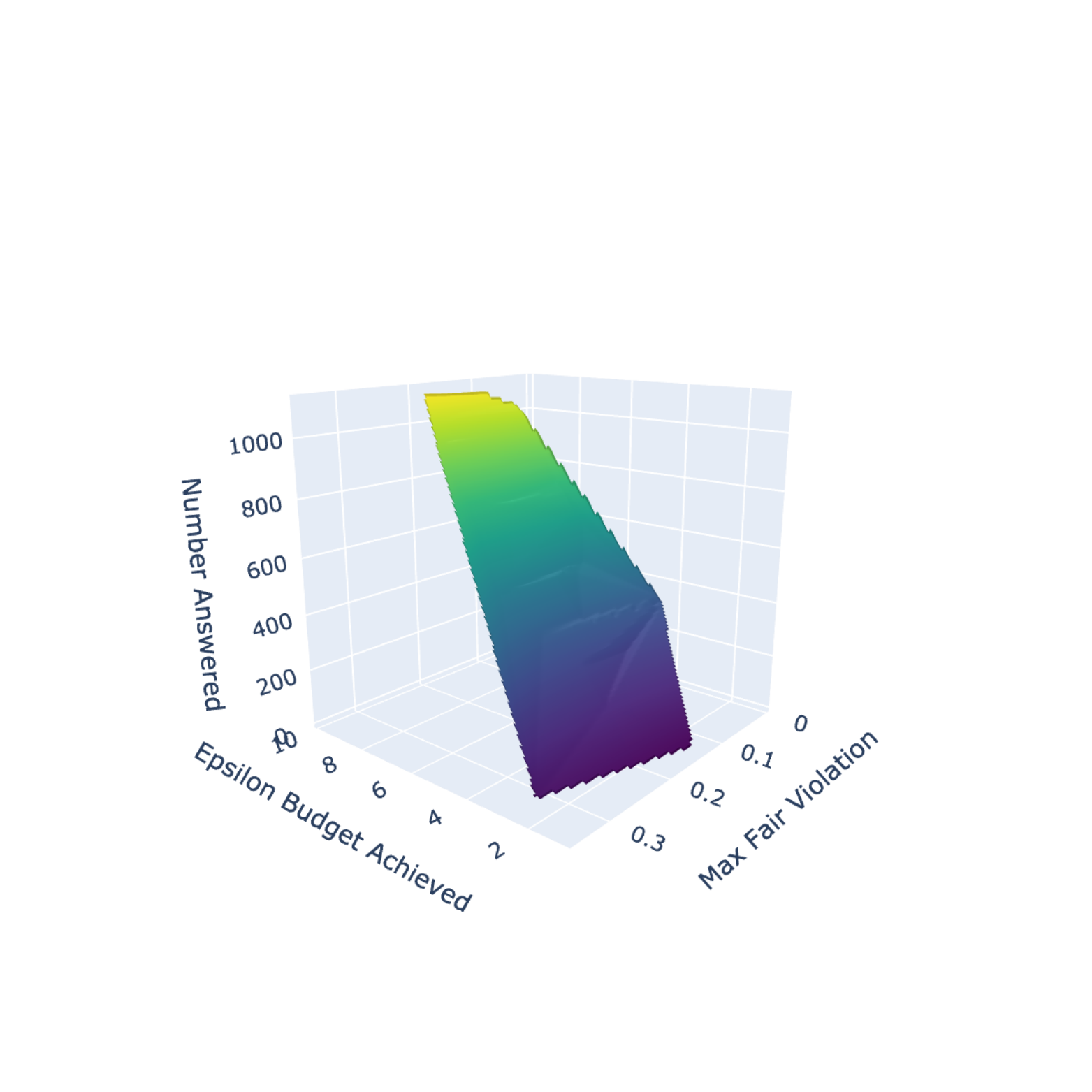}
    \end{subfigure}
    \caption{\textbf{Query Experiment Results on UTKFace}. Experimental setup from~\Cref{tab:datasets}. The left figure shows the the trade-offs between the maximum ensemble query fairness violation $\gamma_{ens}$, the achieved $\varepsilon$, and the number of queries answered. The right figure plots the Pareto-frontier. With increasing privacy budget, more queries can be answered. The same holds when loosening the fairness constraint. At small privacy budgets, small fairness constraint might not be achievable.} 
    \label{fig:utkface results}
\end{figure}

\begin{figure}
\begin{tabular}{lll}
a) ColorMNIST & \includegraphics[width=.3\linewidth]{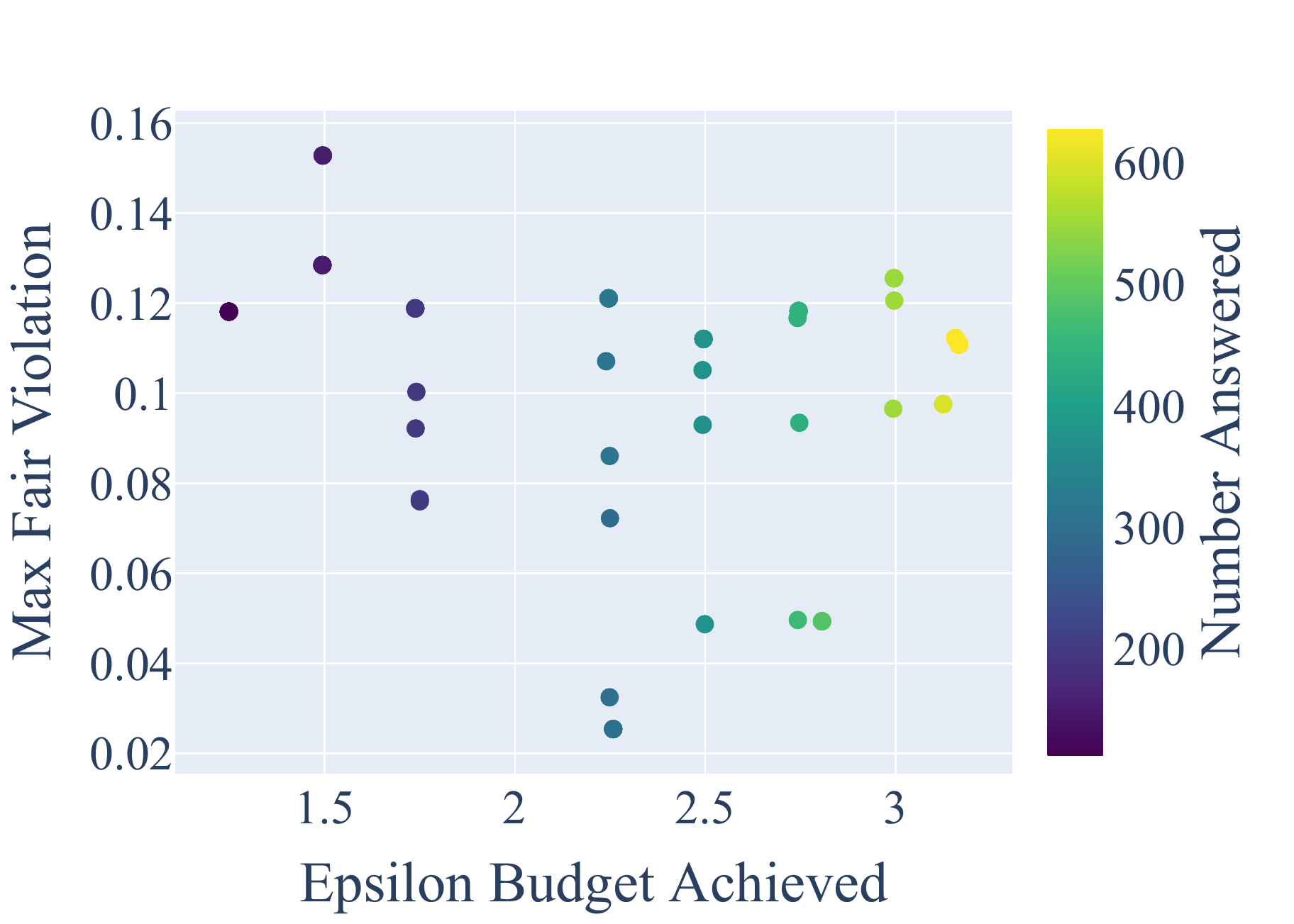} & \includegraphics[width=.3\linewidth, trim=2cm 2cm 3cm 5cm,clip]{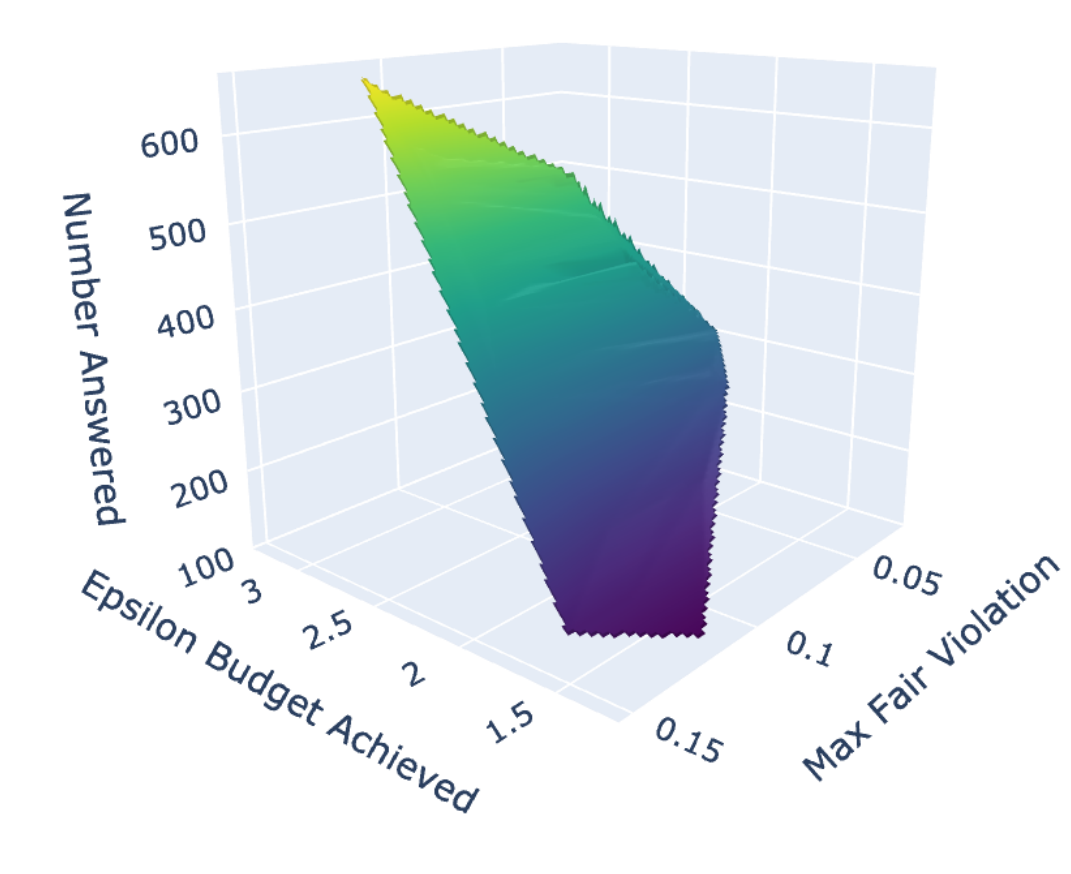} \\
b) CelebA & \includegraphics[width=.3\linewidth]{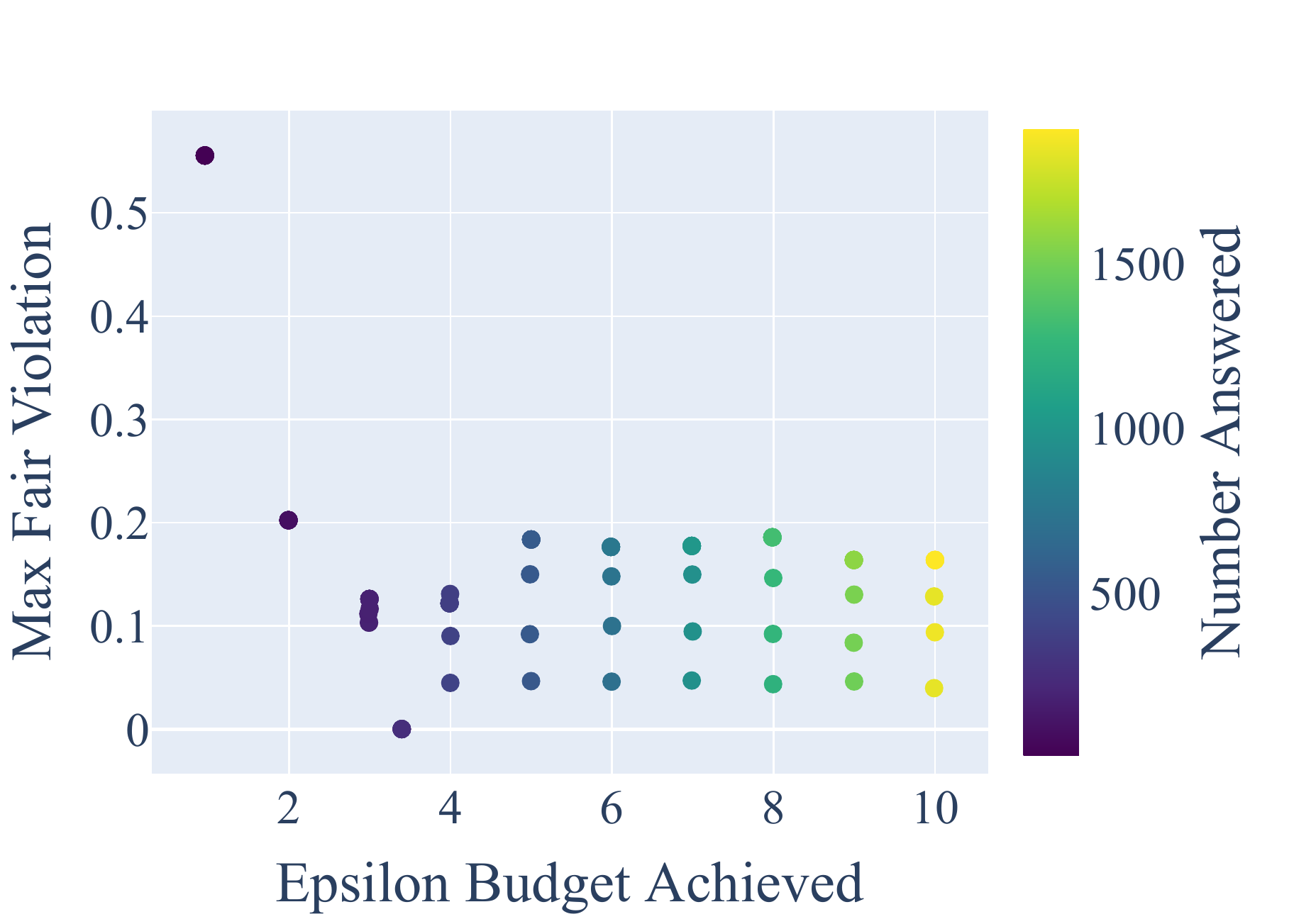} & \includegraphics[width=.3\linewidth,trim=2cm 2cm 3cm 5cm,clip]{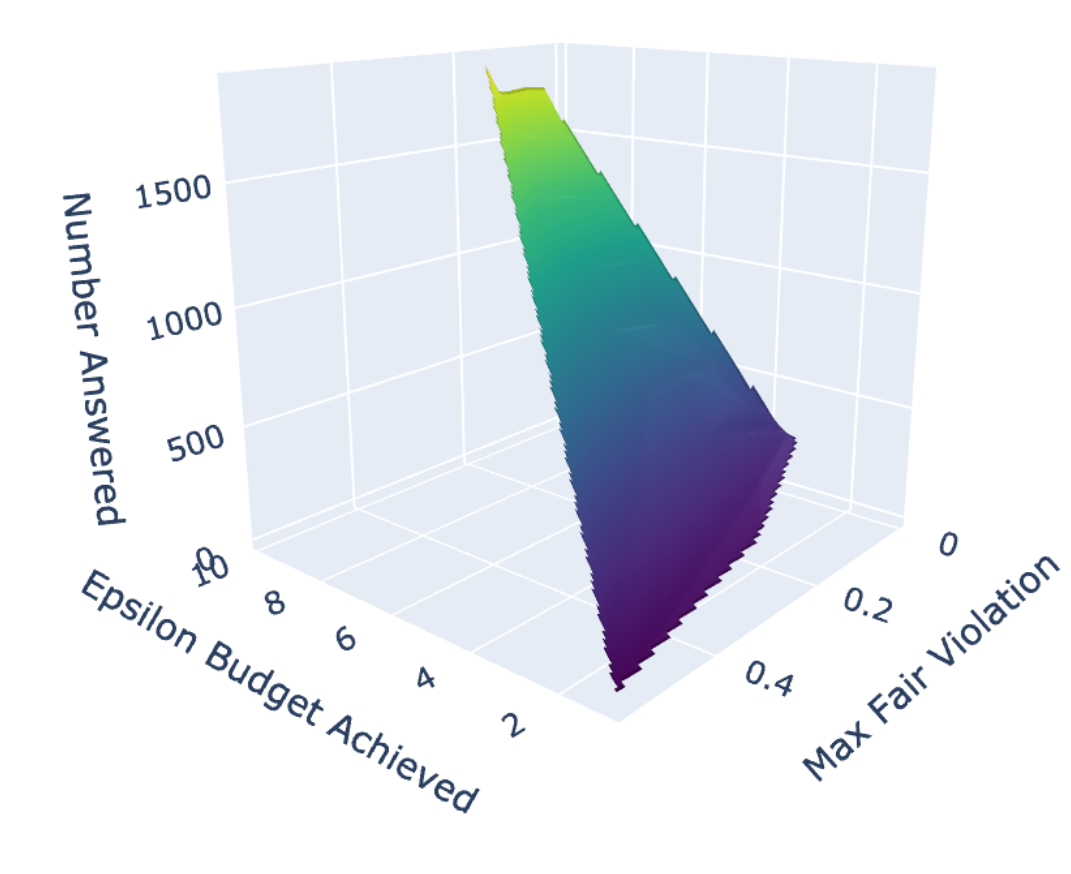} \\
c) FairFace & \includegraphics[width=.3\linewidth]{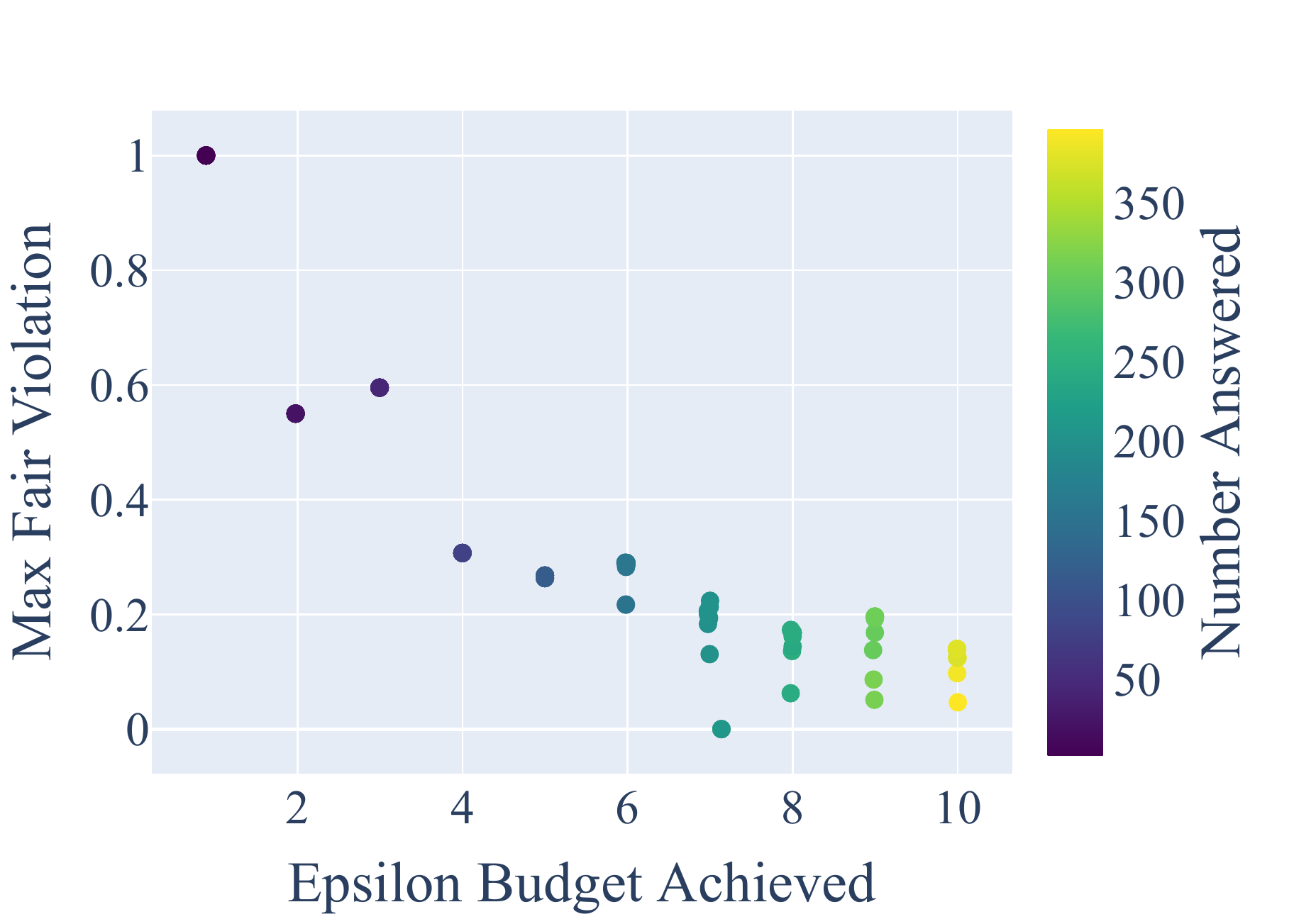} & \includegraphics[width=.3\linewidth,trim=2cm 2cm 3cm 5cm,clip]{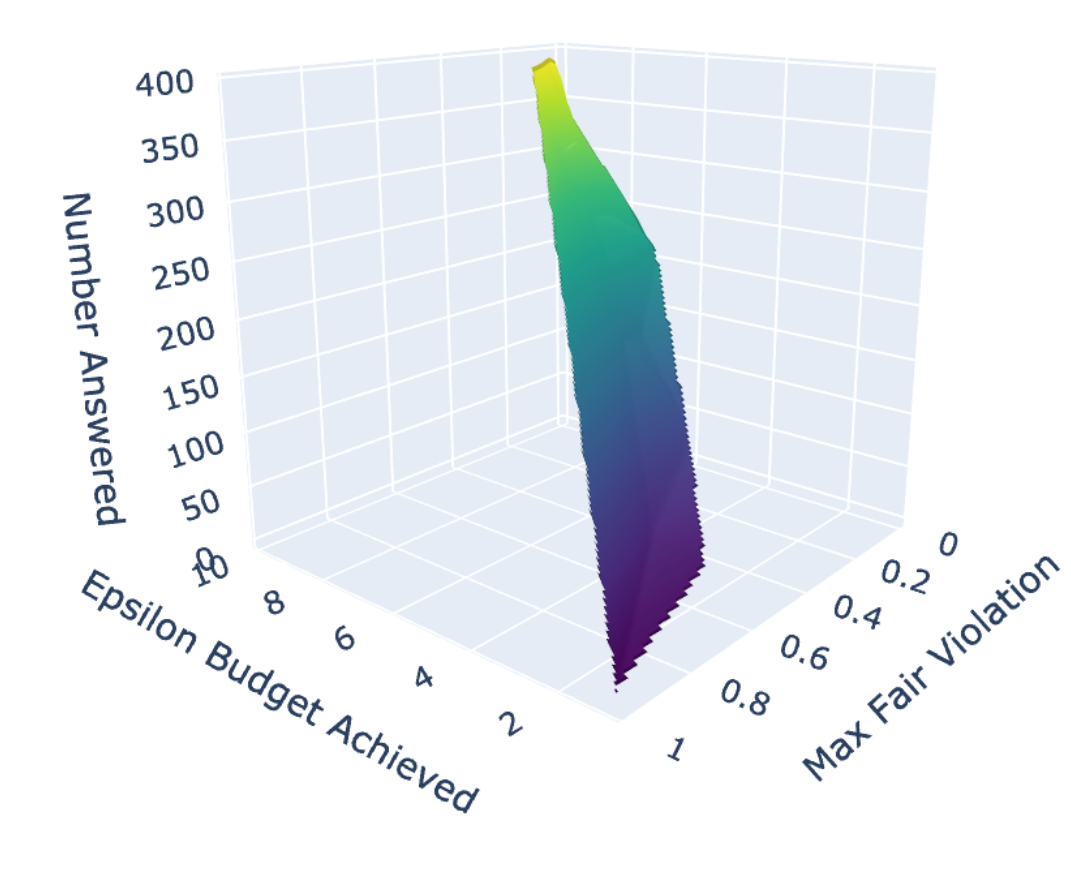} \\
d) CheXpert & \includegraphics[width=.3\linewidth]{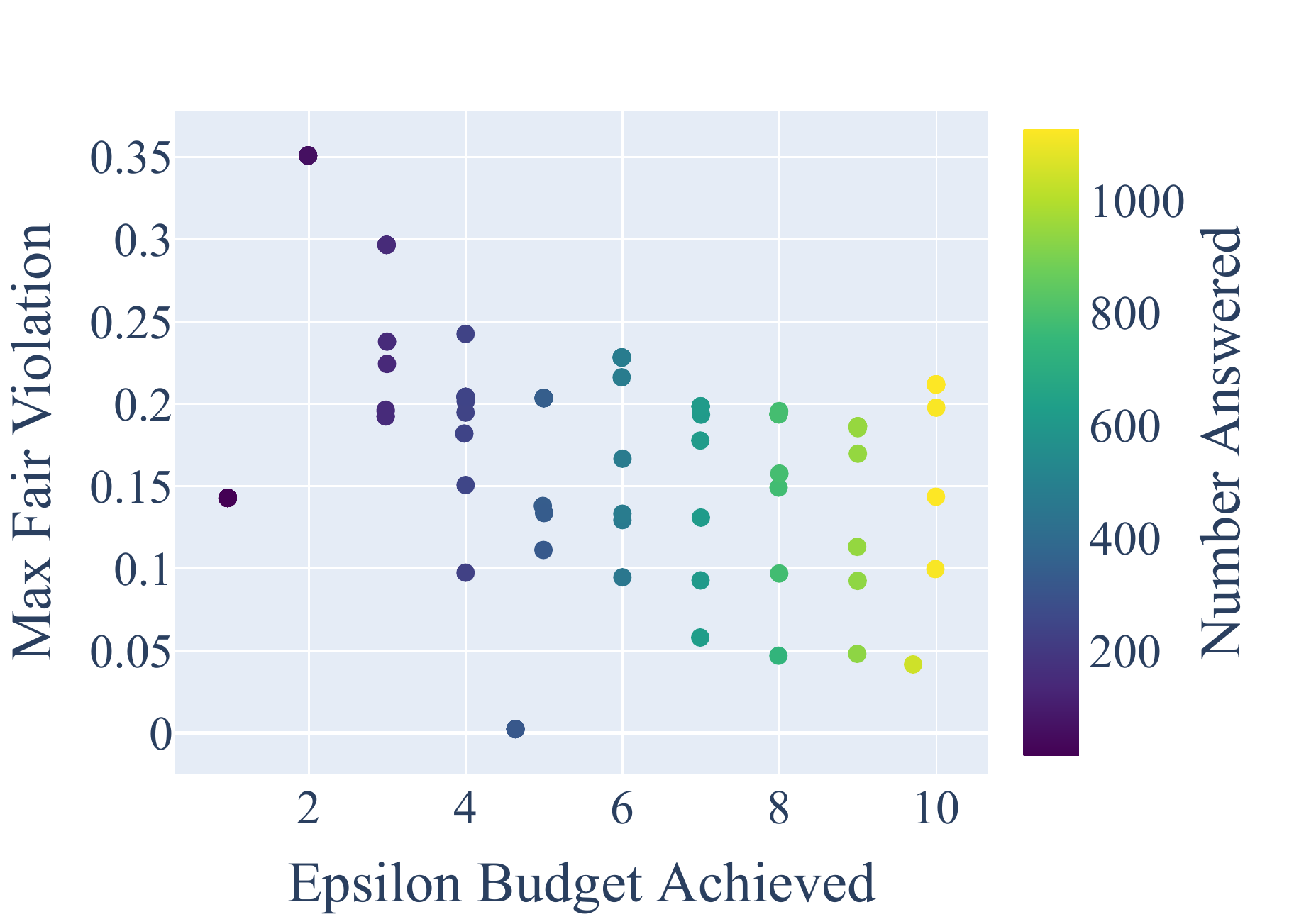} & \includegraphics[width=.3\linewidth,trim=2cm 2cm 3cm 5cm,clip]{images/utkface_pareto_surface} \\
\end{tabular}
\caption{Query experiments on other datasets. Setup described in \Cref{tab:datasets} and discussion is in \Cref{sec:exp-answered-queries}.}
\label{fig:other query results}
\end{figure}

We run a set of query experiments to investigate the trade-offs between privacy, fairness, and model utility. 
We do not train the student model for these querying experiments, but they will be trained later on. Instead, we use the number of queries answered as an estimate of the student model utility since an adequate number of queries needs to be answered to train a student model with good accuracy.%
In the first set of experiments, we run queries with varying consensus threshold $T$ and fairness violation threshold $\rho_{fair}$ at fixed privacy budget $\varepsilon$, and record the number of queries answered. 
We query the teacher ensemble models with varying privacy budget $\varepsilon$ and fairness violation threshold $\rho_{fair}$. For these queries, we measure the maximum fairness violation $\gamma$, the achieved $\varepsilon$, and the number of queries answered. Using these query results, we also select and plot the points on the \pf. %
The results for the UTKFace dataset are shown in \Cref{fig:utkface results}. The results on the other datasets are found in \Cref{fig:other query results}. 

\Cref{fig:utkface results} (left) plots the the trade-offs between the maximum fairness violation $\gamma$, the achieved $\varepsilon$, and the number of queries answered.  As expected, we observe that increasing $\varepsilon$ allows more queries to be answered. Relaxing $\rho_{fair}$ at fixed $\varepsilon$ also leads to more queries being answered, although the effect is not as apparent. Additionally, when $\varepsilon$ is very low, smaller $\gamma$ is not achievable due to having too few queries answered and the fairness regulation mechanism not being activated as a result. 

\Cref{fig:utkface results} (right) plots the \pf of the query results. We plot the privacy constraint, fairness constraint, and the number of queries answered as a 3D plot to better visualize the tension between these different objectives. The figure gives similar insights as the other figure. Another observation is that although smaller $\gamma$ is achievable when a higher number of queries are answered, at some point the fairness constraint needs to be relaxed in order to answer more queries.

\begin{figure}[b]
    \includegraphics[width=0.4\textwidth,trim=2.5cm 3cm 3cm 5cm,clip]{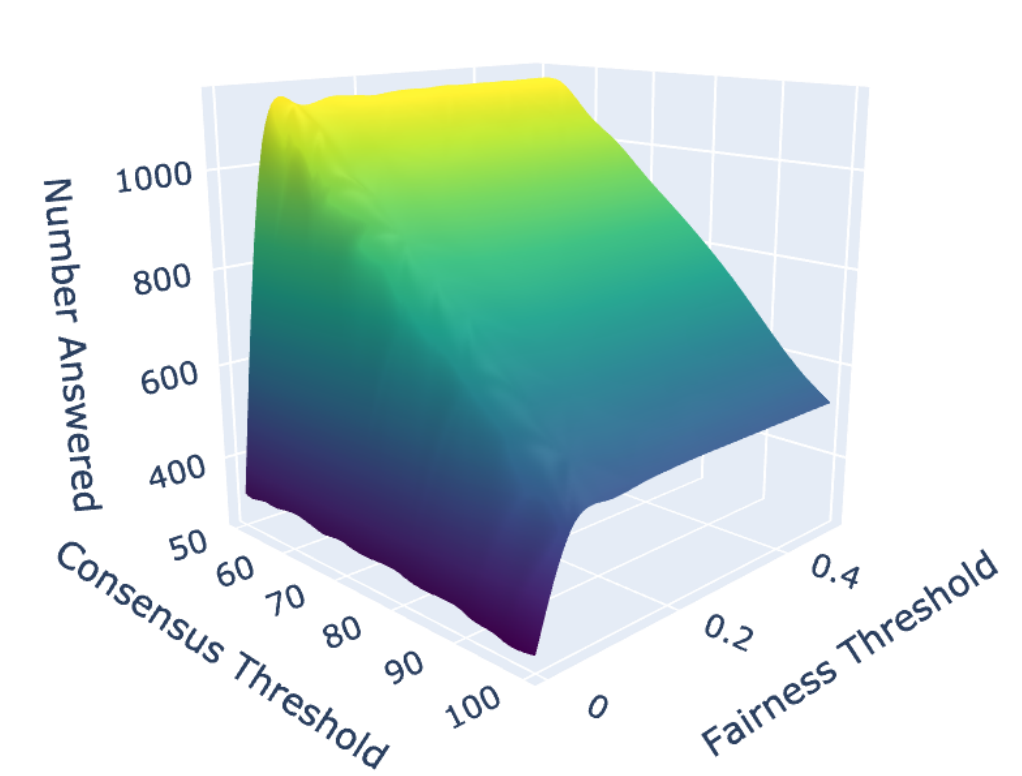}
    \label{fig:query2-utkface} 
    \caption{\textbf{Query Experiment Results on UTKFace}. The figure plots the effect of consensus threshold $T$ and fairness threshold $\gamma_{threshold}$ on the number of queries answered. We observe that the number of queries answered increases with smaller $T$ and larger $\gamma_{threshold}$.}
\end{figure}

\begin{figure}
\centering
    \begin{subfigure}[b]{0.4\textwidth}
        \includegraphics[width=\textwidth,trim=2.5cm 3cm 3cm 5cm,clip]{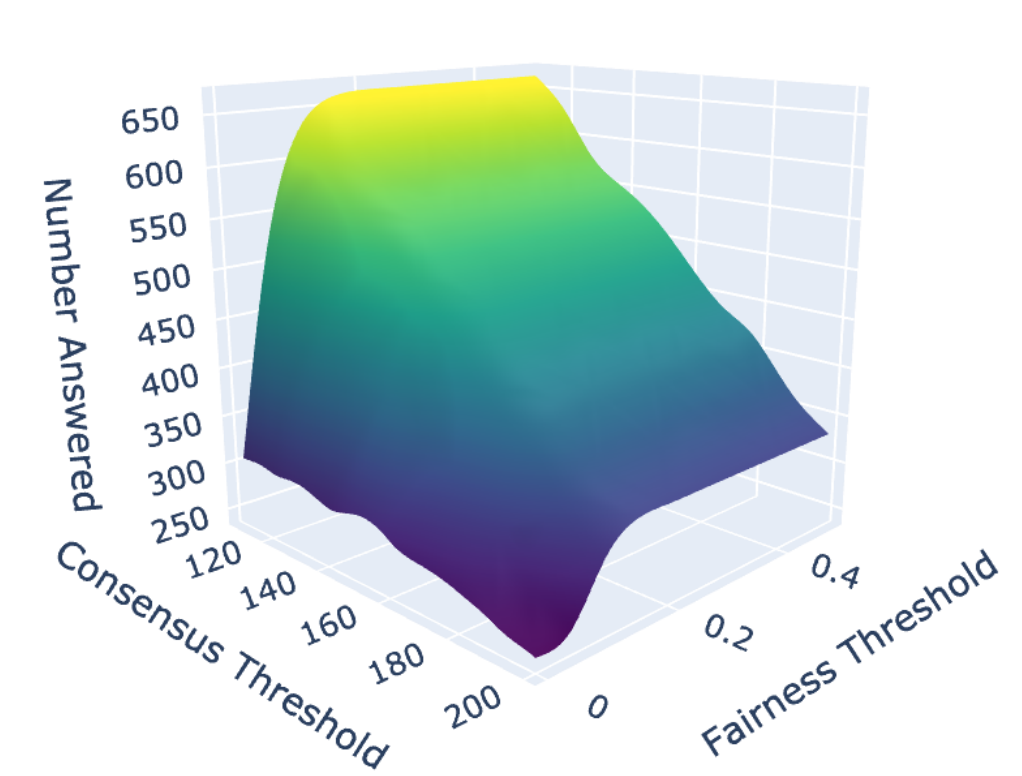}
        \caption{ColorMNIST}
    \end{subfigure}
        \begin{subfigure}[b]{0.4\textwidth}
        \includegraphics[width=\textwidth,trim=2.5cm 3cm 3cm 5cm,clip]{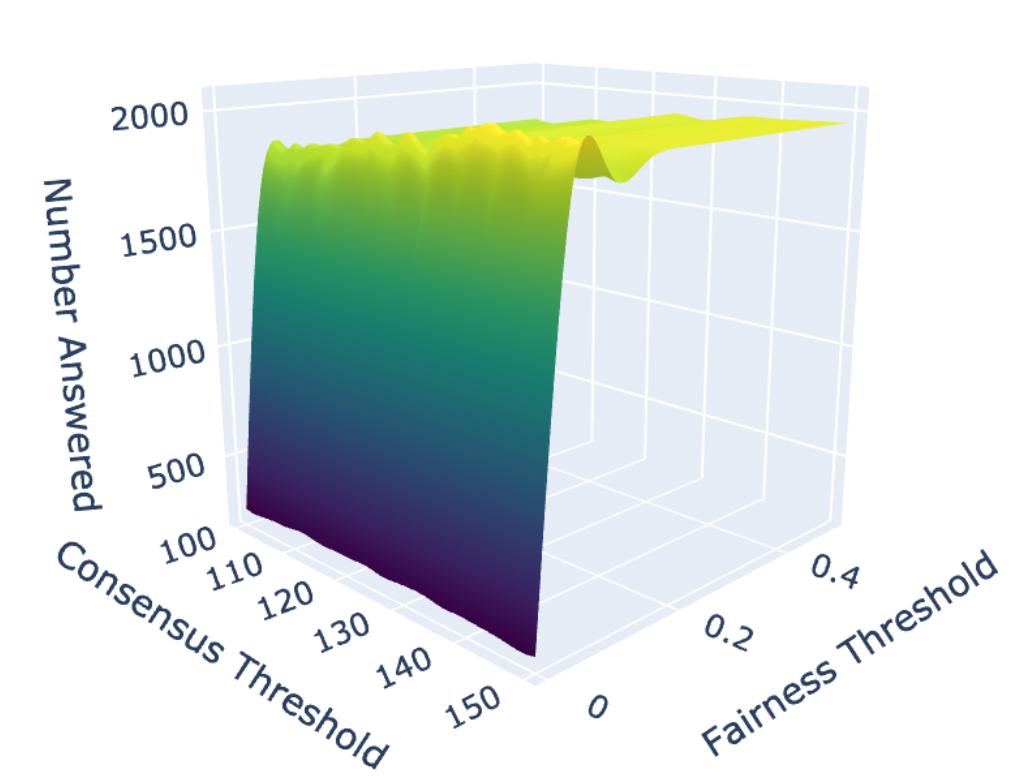}
        \caption{CelebA}
    \end{subfigure}

    \begin{subfigure}[b]{0.4\textwidth}
        \includegraphics[width=\textwidth,trim=2.5cm 3cm 3cm 5cm,clip]{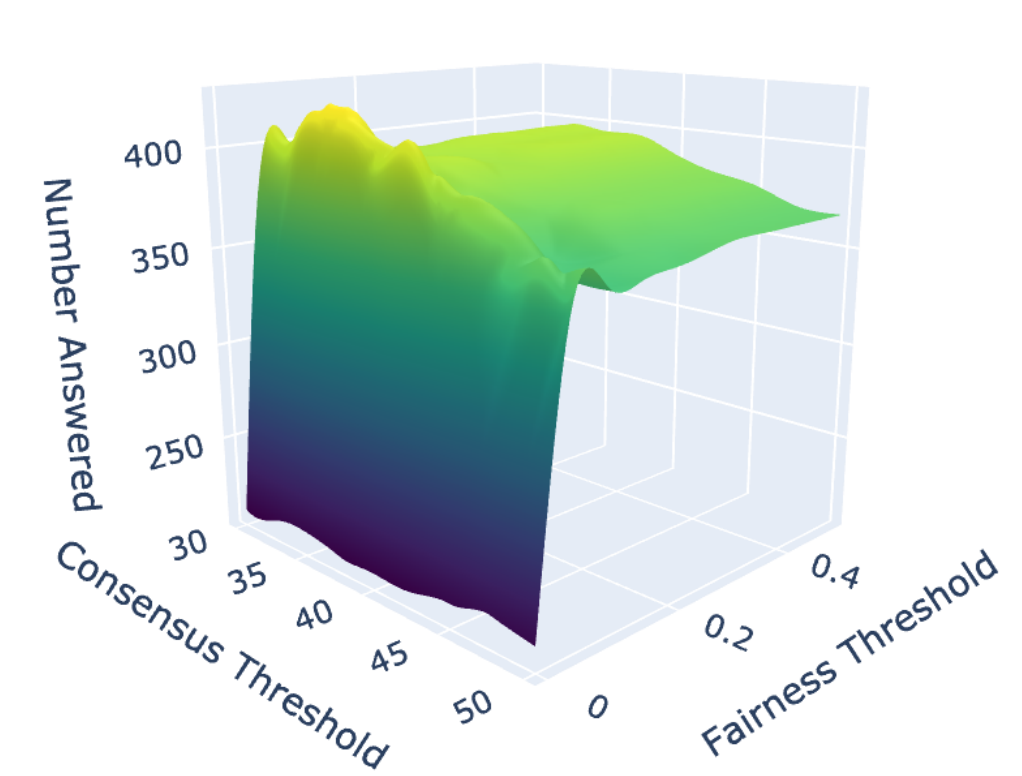}
        \caption{FairFace}
    \end{subfigure}
        \begin{subfigure}[b]{0.4\textwidth}
        \includegraphics[width=\textwidth,trim=2.5cm 3cm 3cm 5cm,clip]{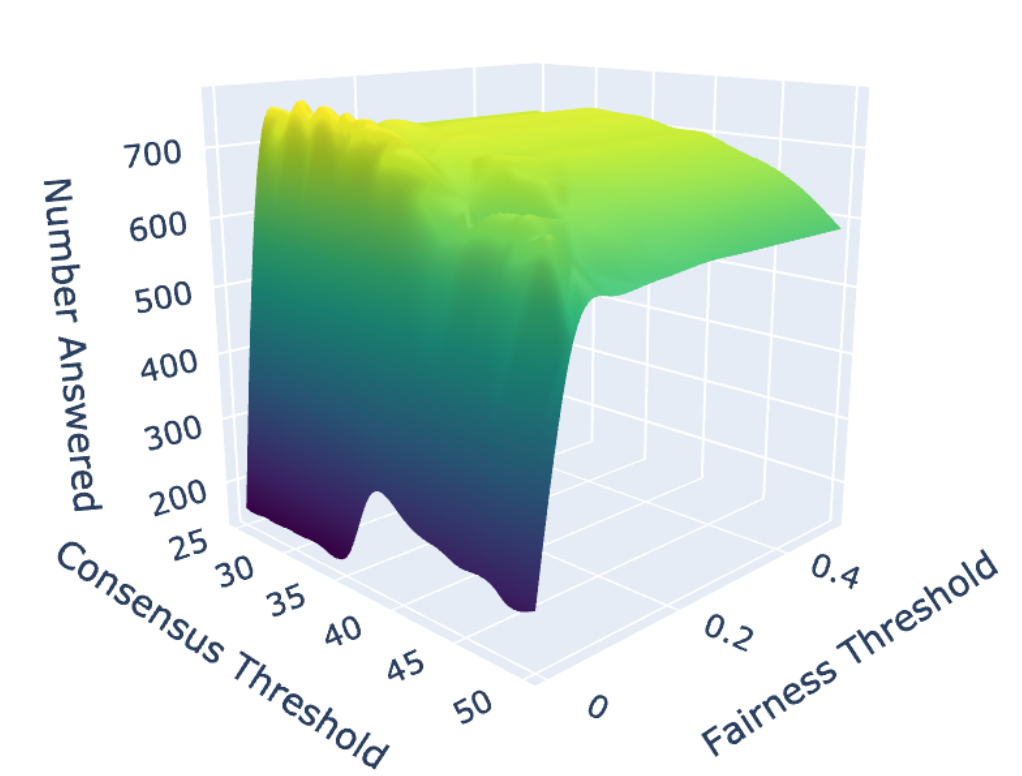}
        \caption{CheXpert}
    \end{subfigure}
\caption{Query experiments on other datasets. Setup described in \Cref{tab:datasets} and discussion is in \Cref{sec:exp-answered-queries}. We found that in order to obtain the best results on student accuracy, some datasets require addition of significant noise $sigma_{1}$, which leads to differences in surfaces' shapes.}
\label{fig:query2}
\end{figure}

We run an additional set of experiments of querying the teacher ensemble models to investigate the effect of different parameters on the number of queries answered. For these experiments, we run queries with varying consensus threshold $T$ and fairness violation threshold $\rho_{fair}$ at fixed privacy budget $\varepsilon$, and record the number of queries answered. \Cref{fig:query2-utkface} plots the results on UTKFace, and the results on other datasets are in \Cref{fig:query2}. The graph shows the effect of varying the consensus threshold $T$ and fairness violation threshold $\rho_{fair}$ on the number of queries answered.  We observe that decreasing $T$ leads to a higher number of queries answered. Similarly, increasing $\rho_{fair}$ to a certain extent also leads to more queries being answered. Once the fairness violation threshold is too large, further relaxing the constraint would not lead to answering more queries, at which point no more queries are rejected due to the fairness constraint. Furthermore, at a fairness constraint of 0, there is a sharp decrease in the number of queries answered. The reason behind this is that if no fairness violation is allowed, no more queries can be answered after the fairness gap reaches 0, as any additional query would break the balance and increase the gap.

\section{Extended Related Work: Integrating Fairness into Private Learning}
In the literature, different fairness notions have been implemented within \dpsgd and PATE frameworks.

\paragraph{Fairness and DP-SGD}
It has been shown that training with \dpsgd leads to disparate accuracy decrease over different data sub-groups~\citep{vinith, farrand2020neither}.
In particular, model accuracy decreases more for underrepresented data from the tails of the distribution~\cite{vinith}.
\citet{farrand2020neither} presented similar findings and observed that privacy can even have a negative impact on the model fairness when the training data is only slightly imbalanced. 
As potential reasons for this, the authors identified the clipping operation in \dpsgd. Since underrepresented data has larger gradients, these gradients are more effected by the clipping operation, and thereby, this data experiences a higher information loss~\cite{farrand2020neither}.
To limit this effect, \citet{xu2021removing} proposed adapting the clipping threshold in \dpsgd individually for each sensitive group.
They showed how their approach limits the disparate impact of \dpsgd on different groups.
However, due to higher information leakage form larger gradients, their method requires larger perturbations.
In a similar vein,~\citet{zhang2021balancing} propose early stopping to mitigate the negative impact of \dpsgd on model fairness.
The authors observe that \dpsgd makes ML model training less stable which they leverage to interrupt training once high-enough fairness is achieved, without a significant loss in accuracy. 
However, all these methods solely manage fairness as an indirect byproduct of adapting the private training mechanism.  Neither of them integrates explicit fairness constraints to yield formal guarantees, such as done in this work.

\citet{tran2021lagrangian} proposed applying a Lagrangian dual approach for solving the joint optimization of fairness and privacy in ML.
Therefore, they rely on a fairness constraint plus adaptive clipping and make the computations of the primal and dual update steps differentially private w.r.t.
the considered sensitive attributes.
However, their method adds a significant computational overhead, especially for larger ML models and mini-batch sizes (increase of up to factor 100).

\paragraph{Fairness and PATE}
When comparing the fairness impact of \dpsgd and PATE, \citet{uniyal2021dp} observed that PATE induces lower accuracy parity. 
The authors reason that this might be because the diversity among the teachers allows to cancel out their individual fairness issues.
However, their observations only hold for very small numbers of teachers (10, in contrast to 250 proposed for MNIST in the original PATE paper~\cite{papernot2016semi}).
This however yields sub-optimal privacy-utility trade-offs since in PATE, stronger privacy guarantees can be obtained when using more teachers which allows for the injection of more noise. 
In the work closest to ours, \citet{tran2021fairness} study fairness properties of PATE and identified both algorithmic properties of the training (number of teachers, regularizer, privacy noise), and properties of the student data (magnitude of the input norm, and distance to the decision boundary)~as factors influencing prediction fairness. 
To mitigate tensions, they proposed releasing the teacher models' prediction histogram as \textit{soft labels} to train the student model.
However, it has been shown that releasing the histograms leaks significant amounts of private information~\cite{wang2022differential}, which makes their method leaks privacy above the promised DP guarantees.
In contrast, in this work, we integrate fairness in the aggregation process while keeping the teachers' votes private, and, thereby providing the promised privacy guarantees.

\end{document}